\documentclass{article}

\usepackage{microtype}
\usepackage{graphicx}
\usepackage{subcaption}
\usepackage{wrapfig}
\usepackage{booktabs} 
\usepackage{multirow}

\usepackage[usenames,dvipsnames]{xcolor}

\usepackage[colorlinks]{hyperref}
\hypersetup{citecolor=MidnightBlue}
\hypersetup{linkcolor=MidnightBlue}
\hypersetup{filecolor=MidnightBlue}
\hypersetup{urlcolor=MidnightBlue}
\usepackage{url}

\newcommand{\mplred}[1]{\textcolor[HTML]{E41A1C}{#1}}
\newcommand{\mplgreen}[1]{\textcolor[HTML]{4DAF4A}{#1}}
\newcommand{\mplblue}[1]{\textcolor[HTML]{377EB8}{#1}}

\usepackage[accepted]{icml2023}

\renewcommand{\icmlEqualContribution}{\textsuperscript{*}Work done while at Secondmind}

\usepackage{amsmath}
\usepackage{amssymb}
\usepackage{mathtools}
\usepackage{amsthm}
\usepackage{nicefrac}
\usepackage{siunitx}

\usepackage{lipsum} 

\usepackage[acronym]{glossaries}
\setacronymstyle{long-sc-short}
\glsdisablehyper

\DeclareRobustCommand{\parhead}[1]{\textbf{#1}~}

\usepackage[capitalize,noabbrev]{cleveref}
\crefname{section}{\S}{\S\S}
\Crefname{section}{\S}{\S\S}
\crefformat{equation}{#2eq.~#1#3}
\Crefformat{equation}{#2Eq.~#1#3}

\usepackage[disable,textsize=tiny]{todonotes}

\newcommand{\g}{\,|\,}

\newcommand{\mbf}{\mathbf{f}}

\newcommand{\mbh}{\mathbf{h}}

\newcommand{\mbk}{\mathbf{k}}

\newcommand{\mbm}{\mathbf{m}}

\newcommand{\mbu}{\mathbf{u}}
\newcommand{\mbv}{\mathbf{v}}
\newcommand{\mbw}{\mathbf{w}}
\newcommand{\mbx}{\mathbf{x}}
\newcommand{\mby}{\mathbf{y}}
\newcommand{\mbz}{\mathbf{z}}

\newcommand{\mbC}{\mathbf{C}}

\newcommand{\mbI}{\mathbf{I}}

\newcommand{\mbW}{\mathbf{W}}
\newcommand{\mbX}{\mathbf{X}}

\newcommand{\mbZ}{\mathbf{Z}}

\newcommand{\mbbeta}{\boldsymbol{\beta}}

\newcommand{\mbeta}{\boldsymbol{\eta}}

\newcommand{\mbmu}{\boldsymbol{\mu}}

\newcommand{\mbxi}{\boldsymbol{\xi}}

\newcommand{\mbSigma}{\boldsymbol{\Sigma}}

\newcommand{\mbzero}{\mathbf{0}}

\DeclareRobustCommand{\KL}[2]{\ensuremath{\textsc{kl}\left[#1\;\|\;#2\right]}}

\newcommand{\cH}{\mathcal{H}}

\newcommand{\cN}{\mathcal{N}}
\newcommand{\cO}{\mathcal{O}}

\newcommand{\cX}{\mathcal{X}}

\newcommand{\bbR}{\mathbb{R}}

\newcommand{\defeq}{\triangleq}
\newcommand{\argdot}{\cdot\,}

\DeclareRobustCommand{\Cov}[2]{\ensuremath{\mathrm{Cov}\left(#1,#2\right)}}

\newcommand{\dummy}{\mbxi}

\newcommand{\featureCoefficient}{\varsigma}
\newcommand{\kernelCoefficient}{\lambda}

\newcommand{\sigmoid}{\sigma}

\newcommand{\Normal}{\cN}
\newcommand{\GP}{\mathcal{GP}}

\newcommand{\variationalLoc}{\mbm}
\newcommand{\variationalCov}{\mbC}

\newcommand{\kernel}{k}
\newcommand{\schurKernel}{s}

\newcommand{\inputDim}{d}
\newcommand{\level}{\ell}

\newcommand{\hiddenLayer}{H}
\DeclareRobustCommand{\sphericalHarmonic}[2]{\ensuremath{{Y}_{#1,#2}}}
\DeclareRobustCommand{\nHarmonics}[2]{\ensuremath{J({#1}, {#2})}}

\newcommand{\shape}{\kappa}

\newcommand{\parallelFunc}{g}

\newcommand{\func}{f}
\newcommand{\funcValues}{\mbf}

\newcommand{\perpFunc}{h}
\newcommand{\perpFuncValues}{\mbh} 

\newcommand{\observedInput}{\mbx}
\newcommand{\observedInputCollection}{\mbX}

\newcommand{\inducingInputCollection}{\mbZ}
\newcommand{\inducingVariables}{\mbu}

\newcommand{\altInducingInput}{\mbw}
\newcommand{\altInducingInputCollection}{\mbW}
\newcommand{\altInducingVariables}{\mbv}

\newcommand{\perpAltInducingVariables}{\mbv'} 

\DeclareRobustCommand{\covarianceMatrix}[3]{\ensuremath{\mathbf{#1}_{#2#3}}}

\DeclareRobustCommand{\K}[2]{\ensuremath{\covarianceMatrix{K}{#1}{#2}}}
\DeclareRobustCommand{\C}[2]{\ensuremath{\covarianceMatrix{S}{#1}{#2}}}
\DeclareRobustCommand{\Q}[2]{\ensuremath{\covarianceMatrix{Q}{#1}{#2}}}

\newcommand{\Kff}{\K{\mbf}{\mbf}}
\newcommand{\Kuu}{\K{\mbu}{\mbu}}
\newcommand{\Kvv}{\K{\mbv}{\mbv}}
\newcommand{\Kfu}{\K{\mbf}{\mbu}}
\newcommand{\Kuf}{\K{\mbu}{\mbf}}
\newcommand{\Kvf}{\K{\mbv}{\mbf}}
\newcommand{\Kvu}{\K{\mbv}{\mbu}}

\newcommand{\Cff}{\C{\mbf}{\mbf}}
\newcommand{\Cvv}{\C{\mbv}{\mbv}}
\newcommand{\Cvf}{\C{\mbv}{\mbf}}
\newcommand{\Cfv}{\C{\mbf}{\mbv}}

\newcommand{\Qff}{\Q{\mbf}{\mbf}}
\newcommand{\Qfu}{\Q{\mbf}{\mbu}}
\newcommand{\Quf}{\Q{\mbu}{\mbf}}
\newcommand{\Qvf}{\Q{\mbv}{\mbf}}
\newcommand{\Qvv}{\Q{\mbv}{\mbv}}
\newcommand{\Quv}{\Q{\mbu}{\mbv}}
\newcommand{\Qvu}{\Q{\mbv}{\mbu}}

\newcommand{\Matern}{Mat\'{e}rn}
\newcommand{\Nystroem}{Nystr\"{o}m}

\newacronym[longplural={Gaussian processes}]{GP}{gp}{Gaussian process}
\newacronym[longplural={deep \textsc{gp}s}]{DGP}{dgp}{deep \textsc{gp}}
\newacronym{SVGP}{svgp}{sparse variational \textsc{gp}}
\newacronym{ODVGP}{odvgp}{orthogonally-decoupled variational \textsc{gp}}
\newacronym{SOLVEGP}{solvegp}{\underline{s}parse \underline{o}rthogona\underline{l} \underline{v}ariational inf\underline{e}rence for \underline{\textsc{gp}}s}

\newacronym{MLE}{mle}{maximum likelihood estimation}
\newacronym{MAP}{map}{maximum \textit{a posteriori}}
\newacronym{MCMC}{mcmc}{Markov chain Monte Carlo}
\newacronym{MC}{mc}{Monte Carlo}
\newacronym{VI}{vi}{variational inference}

\newacronym{VFF}{vff}{variational Fourier feature}

\newacronym{CI}{ci}{confidence interval}

\newacronym{NN}{nn}{neural network}
\newacronym{DNN}{dnn}{deep neural network}
\newacronym{BNN}{bnn}{Bayesian neural network}
\newacronym{SE}{se}{squared exponential}

\newacronym{KL}{kl}{Kullback–Leibler}
\newacronym{RKHS}{rkhs}{reproducing kernel Hilbert space}

\newacronym{RELU}{relu}{rectified linear unit}
\newacronym{SOFTPLUS}{softplus}{Softplus}

\newacronym{UCI}{uci}{UC Irvine}

\newacronym{NLPD}{nlpd}{negative log predictive density}
\newacronym{RMSE}{rmse}{root-mean-square error}

\newacronym{ELBO}{elbo}{evidence lower bound}

\newacronym{LBFGS}{l-bfgs}{low-memory Broyden–Fletcher–Goldfarb–Shanno}

\icmltitlerunning{Spherical Inducing Features for Orthogonally-Decoupled Gaussian Processes}

\begin{document}

\twocolumn[
\icmltitle{Spherical Inducing Features for Orthogonally-Decoupled Gaussian Processes}



\icmlsetsymbol{equal}{*}

\begin{icmlauthorlist}
\icmlauthor{Louis C.~Tiao}{usyd,equal}
\icmlauthor{Vincent~Dutordoir}{ucam,sm}
\icmlauthor{Victor~Picheny}{sm}
\end{icmlauthorlist}

\icmlaffiliation{usyd}{University of Sydney, Sydney, Australia}
\icmlaffiliation{ucam}{University of Cambridge, Cambridge, UK}
\icmlaffiliation{sm}{Secondmind, Cambridge, UK}

\icmlcorrespondingauthor{L. Tiao}{louis.tiao@sydney.edu.au}

\icmlkeywords{Machine Learning, Gaussian Processes, Variational Inference, ICML}

\vskip 0.3in
]



\printAffiliationsAndNotice{\icmlEqualContribution} 

\todo{Open issue \#1: we're overloading notation $k$ for kernel and the index for orthogonal inducing variables}

\begin{abstract}
Despite their many desirable properties, \glspl{GP} are often compared 
unfavorably to deep \glspl{NN}  for lacking the ability to learn representations. 
Recent efforts to bridge the gap between \glspl{GP} and deep \glspl{NN} have 
yielded a new class of inter-domain variational \glspl{GP} in which the 
inducing variables correspond to hidden units of a feedforward \gls{NN}. 
In this work, we examine some practical issues associated with this approach 
and propose an extension that leverages the orthogonal decomposition 
of \glspl{GP} to mitigate these limitations. 
In particular, we introduce spherical inter-domain features to construct more 
flexible data-dependent basis functions for both the principal and orthogonal 
components of the \gls{GP} approximation and show that incorporating \gls{NN} 
activation features under this framework not only alleviates these shortcomings
but is more scalable than alternative strategies.
Experiments on multiple benchmark datasets demonstrate the effectiveness of 
our approach. 
\end{abstract}

\glsunset{SOFTPLUS}
\glsunset{SOLVEGP}

\glsreset{GP}
\glsreset{NN}

\section{Introduction}
\label{sec:introduction}



\Glspl{GP}
provide a powerful framework for reasoning about unknown functions
and are ubiquitous in 
probabilistic 
machine learning~\cite{rasmussen2006gaussian}.
They
are data-efficient, 
innately 
robust to over-fitting, 
and can flexibly encode priors through their covariance function.
Last but not least, by virtue of their ability to 
faithfully
capture predictive uncertainty, they 
form the backbone of many
sequential decision-making methods that require reliable uncertainty estimates 
to balance trade-offs such as exploration and exploitation, e.g. in 
reinforcement learning~\cite{deisenroth2011pilco}, 
Bayesian optimization~\cite{garnett_bayesoptbook_2023}, 
and probabilistic numerics~\cite{hennig2022probabilistic}.
In spite of
their many advantages, \glspl{GP} are often compared unfavourably to 
deep \glspl{NN} 
for their 
poor scalability to large datasets, and their
inability 
to capture rich hierarchies of abstract representations~\cite{calandra2016manifold,wilson2016deep,ober2021promises}.
\todo{WIP (second pass); up to here}
\todo{non-Gaussian observations is another limitation but not really worth mentioning here}
While \glspl{GP} are the infinite-width limit of \glspl{NN} and therefore, 
in theory, have infinitely more basis functions~\cite{neal1996bayesian}, 
these basis functions 
are 
static
and 
fully determined 
by 
the covariance function~\cite{mackay1998introduction}. 
This makes it difficult for \glspl{GP} to flexibly adapt to 
complex and structured data
from which it is beneficial for the basis functions to learn and encode 
useful
representations.
\todo{citation needed}



Considerable research effort has been devoted to sparse approximations for \glspl{GP}
~\cite{csato2002sparse,seeger2003fast,quinonero2005unifying,snelson2005sparse}.
Not least of these is \glspl{SVGP}~\cite{titsias2009variational,hensman2013gaussian,hensman2015scalable}.
Such advances have not only improved the scalability of \glspl{GP},
but also unlocked more flexibility in model 
specification.
In particular, the use of 
\emph{inter-domain} inducing variables 
in \gls{SVGP}~\citep{lazaro2009inter}
effectively 
equips
the \gls{GP} approximation
with data-dependent basis functions.
\todo{maybe cite convolutional GPs (van der wilk) as an example}
Recent works have 
exploited
this 
to construct 
a new 
family
of \gls{SVGP} models in which the
basis functions 
correspond to 
activations 
of a feed-forward \gls{NN}~\citep{sun2020neural,dutordoir2021deep}.
By stacking multiple layers
to form a \gls{DGP}~\citep{damianou2013deep}, 
the propagation of the predictive 
distribution
accurately resembles
a forward-pass through a deep \gls{NN}.
\todo{remark that this achieves SOTA results, and has increased the representation capacity}

In this paper, we show that 
while this approach results in a posterior predictive 
with a more expressive mean,
the 
variance estimate
is typically less accurate and tends to be over-dispersed.
Additionally, we examine some practical challenges associated with this method, 
such as limitations on the use of certain popular kernel and \gls{NN} 
activation choices. 
To address these issues, we propose an extension that aims to 
mitigate
these limitations.
Specifically, when viewed from the function-space perspective, the posterior 
predictive of \gls{SVGP}
depends
on a single set of basis functions that is 
determined by
only a finite collection 
of inducing variables. 
Recent advances introduce an orthogonal set of basis functions as 
a means of 
capturing
additional variations remaining from the 
standard
basis~\cite{salimbeni2018orthogonally,cheng2017variational,shi2020sparse}. 
We extend this framework by 
introducing 
inter-domain variables 
to construct more 
flexible
data-dependent
basis functions for both the standard and orthogonal components. 
In particular, we show that incorporating 
\gls{NN}
activation 
inducing 
functions
under this framework 
is an effective way to
ameliorate the aforementioned shortcomings. 
Our 
experiments 
on numerous benchmark datasets
demonstrate
that this extension leads to improvements in 
predictive performance against comparable alternatives.





\glsreset{GP}

\section{Background}
\label{sec:background}

\Glspl{GP} are a flexible class of distributions over functions.
A random function $f: \cX \to \bbR$ 
on some domain $\cX \subseteq \bbR^{\inputDim}$ 
is 
distributed according to
a \gls{GP} if, at any finite collection of 
input locations $\mbX_\ast \subseteq \cX$, 
its values $\funcValues_\ast = f(\mbX_\ast)$ follow a 
Gaussian distribution.
A \gls{GP} is fully determined by its mean function, which can be assumed 
without loss of generality 
to be constant (e.g. zero), and its covariance function $\kernel(\observedInput, \observedInput')$.

Consider a supervised learning problem in which we have 
a 
dataset $\{ \observedInput_n, y_n \}_{n=1}^N$ consisting of 
scalar outputs $y_n$, which are 
related 
to $f_n \defeq f(\observedInput_n)$, the value of some 
unknown
function $f(\cdot)$ at input $\observedInput_n \in \cX$,
through the 
likelihood $p(y_n \g f_n, \theta)$.
A powerful modelling approach consists of specifying a \gls{GP} prior on 
the latent function~$f(\cdot)$,
\begin{equation} \label{eq:gaussian-process-prior}
  f(\observedInput) \sim \GP\left(0, \kernel(\observedInput, \observedInput')\right).
\end{equation}
Let $\mbX$ 
denote the inputs, $\funcValues$ the corresponding latent function values, 
and $\mby$ 
the 
outputs.
In the regression setting, the outputs $\mby$ are noisy observations of 
the latent 
values $\funcValues$, typically related through a Gaussian 
likelihood $p(\mby \g \funcValues, \theta) = \Normal(\mby \g \funcValues, \beta^{-1} \mbI)$ 
for some precision $\beta > 0$.
In this 
case, 
although the posterior $p(\funcValues \g \mby)$ is analytically tractable, its
computation
has time complexity $\cO(N^3)$.  

\glsreset{SVGP}

\subsection{Sparse Gaussian processes}
\label{sub:sparse_gaussian_processes}

A range of 
sparse \gls{GP} methods 
have been developed over the years to mitigate these 
limitations~\citep{csato2002sparse,seeger2003fast,quinonero2005unifying,snelson2005sparse}.
Broadly, in sparse \glspl{GP}, one 
summarizes $f(\cdot)$ succinctly in terms 
of \emph{inducing variables}, 
which are 
values $\inducingVariables \defeq f(\mbZ)$ taken at a collection of $M$ (usually $M \ll N$)
locations $\mbZ = [\mbz_1 \cdots \mbz_M]^\top$, where $\mbz_m \in \cX$.
Not least among these approaches is \gls{SVGP}, which casts sparse \glspl{GP} within the 
framework of \gls{VI}~\citep{titsias2009variational,hensman2013gaussian,hensman2015scalable}.

Specifically, the joint distribution of the model augmented by inducing 
variables $\inducingVariables$ is $p(\mby, \funcValues, \inducingVariables) = p(\mby \g \funcValues) p(\funcValues, \inducingVariables)$ 
where 
$p(\funcValues, \inducingVariables) = p(\funcValues \g \inducingVariables) p(\inducingVariables)$ for prior $p(\inducingVariables) = \Normal(\mbzero, \Kuu)$ 
and conditional 
\begin{equation} \label{eq:prior-conditional}
  p(\funcValues \g \inducingVariables) = \Normal(\funcValues \g \Qfu \inducingVariables, \Kff - \Qff),
\end{equation}
where $\Qff \defeq \Qfu \Kuu \Quf$ and $\Qfu \defeq \Kfu \Kuu^{-1}$.  
The joint variational distribution is defined 
as $q(\funcValues, \inducingVariables) \defeq p(\funcValues \g \inducingVariables) q(\inducingVariables)$, 
where $q(\inducingVariables) \defeq \cN\left(\variationalLoc_{\inducingVariables}, \variationalCov_{\inducingVariables} \right)$ 
for variational parameters $\variationalLoc_{\inducingVariables} \in \bbR^M$ 
and $\variationalCov_{\inducingVariables} \in \bbR^{M \times M}$ s.t. $\variationalCov_{\inducingVariables} \succeq 0$. \todo{Splitting hairs here, but positive-definiteness does not imply symmetry}
Integrating out $\inducingVariables$ yields the
posterior
predictive 
\begin{equation*} \label{eq:variational-marginal-f}
  q(\funcValues_\ast) 
  = \Normal \left (\Q{\ast}{\inducingVariables} \variationalLoc_{\inducingVariables}, 
    \K{\ast}{\ast} - \Q{\ast}{\inducingVariables} (\Kuu - \variationalCov_{\inducingVariables}) \Q{\inducingVariables}{\ast} \right ),
\end{equation*}
where parameters $\variationalLoc_{\inducingVariables}$ and $\variationalCov_{\inducingVariables}$ are learned
by minimizing the \gls{KL} divergence between the approximate and exact 
posterior,~$\KL{q(\funcValues)}{p(\funcValues \g \mby)}$.
\todo{The subscript $\ast$ notation was never explicitly introduced.}
Thus seen, 
\gls{SVGP} has time complexity $\cO(M^3)$ at prediction time
and $\cO(M^3 + M^2 N)$ during training.

In the \gls{RKHS} associated
with $k$,
the predictive has a dual representation in which the mean and 
covariance share the same basis 
determined by~$\mbu$~\cite{cheng2017variational,salimbeni2018orthogonally}.
More specifically, the basis function is effectively the vector-valued 
function  $\mbk_\inducingVariables: \cX \to \bbR^M$ 
whose $m$-th component is defined as 
\begin{equation} \label{eq:cross-covariance-vector}
  \left[\mbk_\inducingVariables(\observedInput)\right]_m \defeq \Cov{f(\observedInput)}{u_m}.
\end{equation}
In the 
standard
definition of inducing points, $\left[\mbk_\inducingVariables(\observedInput)\right]_m = k(\mbz_m, \observedInput)$, 
so the basis function is solely determined by 
$k$ and the local influence of pseudo-input $\mbz_m$.

\emph{Inter-domain} inducing features are a 
generalization of 
standard inducing 
variables 
in which 
each variable 
$u_m \defeq L_m[f]$
for some linear operator $L_m: \bbR^\cX \to \bbR$.
A particularly useful operator is the integral transform,
$L_m[f] \defeq \int_\cX f(\observedInput) \phi_m(\observedInput) \, \mathrm{d}\observedInput$,
which was originally 
employed
by \citet{lazaro2009inter}.
Refer to the manuscript of \citet{van2020framework} for a more thorough and 
contemporary treatment.
A closely related form is the 
scalar projection of $f$ onto 
some $\phi_m$ in the \gls{RKHS} $\cH$, 
\begin{equation} \label{eq:projection}
  L_m[f] \defeq \langle f, \phi_m \rangle_\cH,
\end{equation}
which leads to $\left[\mbk_\inducingVariables(\observedInput)\right]_m = \phi_m(\mbx)$ by the reproducing property of the \gls{RKHS}.
This, in effect, equips the \gls{GP} approximation with 
basis functions $\phi_m$ that are not solely determined by the kernel, 
and suitable choices can lead to sparser representations and considerable
computational benefits~\citep{JMLR:v18:16-579,burt2020variational,dutordoir2020sparse,sun2021scalable}.

\subsection{Spherical Harmonics Inducing Features}

An instance of inter-domain features in the form of \cref{eq:projection} 
are the \glspl{VFF}~\citep{JMLR:v18:16-579}, in which $\phi_m$ 
form an orthogonal basis of trigonometric functions.
This formulation offers significant computational advantages but scales poorly 
beyond 
a small handful of 
dimensions.
To address this,~\citet{dutordoir2020sparse} propose a generalization 
of \glspl{VFF} using the spherical harmonics for $\phi_m$, 
which can be viewed as a multi-dimensional extension of the Fourier basis.

The construction relies on the Mercer decomposition of \emph{zonal} kernels, which
can be seen as the analog of stationary kernels in Euclidean spaces,
but for hyperspheres.
They can be expressed 
as $\kernel(\observedInput, \observedInput') = \shape \left(\tilde{\observedInput}^\top \tilde{\observedInput}'\right)$ 
for some \emph{shape function} $\shape: [-1, 1] \to \bbR$, 
where $\tilde{\mbeta} \defeq \frac{\mbeta}{\lVert \mbeta \rVert} \in \mathbb{S}^{\inputDim-1}$ 
for any $\mbeta \in \bbR^{\inputDim}$ .
Loosely speaking, just as stationary kernels are determined by 
the \emph{distance} between inputs, zonal kernels depend only on 
the \emph{angle} between inputs.

The spherical harmonics form an orthonormal basis on $L_2(\mathbb{S}^{\inputDim-1})$ consisting of the
eigenfunctions of the kernel operator $\mathcal{K}$: $\mathcal{K} \sphericalHarmonic{\level}{j} = a_{\level} \sphericalHarmonic{\level}{j}$,
where $\sphericalHarmonic{\level}{j}$ is the spherical harmonic 
of level $\level$ and 
order 
$j$,
and $a_{\level}$ is the corresponding eigenvalue, or Fourier coefficient.
Conveniently,
by the Funk-Hecke theorem,
$a_{\level}$ can be 
computed by the one-dimensional integral
\begin{equation*}
  a_{\level} = \frac{\Omega_\inputDim}{C_{\level}^{(\alpha)}(1)} \int_{-1}^{1} \shape(t) C_{\level}^{(\alpha)}(t) (1-t^2)^{\frac{\inputDim-3}{2}} \, \mathrm{d}t,
\end{equation*}
where $C_{\level}^{(\alpha)}$ is the Gegenbauer polynomial of degree $\level$ and
$\alpha \defeq \frac{(\inputDim-1)}{2}$.
Now, the number $\nHarmonics{\inputDim}{\level}$ of spherical harmonics that exist at a given level $\level$ is determined by the multiplicity of eigenvalue $a_{\level}$.
\todo{just realized I forgot to define $\Omega_\inputDim$}
Thus, $\shape(t)$ can be represented by
\begin{equation} \label{eq:mercer}
  \shape(t) = 
    \lVert \dummy \rVert \lVert \dummy' \rVert 
    \sum_{\level=0}^{\infty} \sum_{j=1}^{\nHarmonics{\inputDim}{\level}} 
      a_{\level}
      \sphericalHarmonic{\level}{j}(\tilde{\dummy}) 
      \sphericalHarmonic{\level}{j}(\tilde{\dummy}'),
\end{equation}
where $t \defeq \tilde{\dummy}^\top \tilde{\dummy}'$ for $\dummy,\dummy' \in \mathbb{R}^\inputDim$.
Refer to the manuscript (Appendix B) of \citet{dutordoir2021deep} for a concise 
summary of
spherical harmonics in multiple dimensions.

Importantly, \cref{eq:mercer} directly yields a Mercer decomposition for 
zonal kernels.
In particular,
let $\kernelCoefficient_{\level}$ denote the Fourier coefficients 
associated with
kernel $k$.
This
gives rise to
the inter-domain 
features
$\phi_{m} \defeq \sphericalHarmonic{\level}{j}$,
where $m$ indexes the pairs $(\level, j)$.
Crucially, this leads to a diagonal covariance 
\begin{equation*}
  \left[\Kuu\right]_{mm'} \defeq \Cov{u_{m}}{u_{m'}} = \kernelCoefficient_m^{-1} \delta_{mm'},
\end{equation*}
where $\kernelCoefficient_m \defeq \kernelCoefficient_{\level}$ 
and $\delta$ denotes the Kronecker delta.

\subsection{Neural Network Inducing Features}
\label{sub:neural_network_inducing_features}

The 
recent
works of \citet{sun2020neural,dutordoir2021deep} aim to 
construct inter-domain features $\phi_m$ such 
that $\mbk_\inducingVariables(\observedInput)$ 
in \cref{eq:cross-covariance-vector} 
corresponds to a
hidden
layer in a feed-forward \gls{NN}: $\sigmoid(\mbbeta \observedInput)$,
for some $\mbbeta \in \bbR^{M \times \inputDim}$ and 
activation
$\sigmoid$ such as 
the \gls{SOFTPLUS} or the \gls{RELU} function.

In particular, let $\hiddenLayer_m: \cX \to \bbR$ denote the output of 
the $m$-th hidden unit. 
Additionally, let us project this function onto the unit hypersphere,
\begin{equation} \label{eq:spherical-activation-feature}
  \hiddenLayer_m(\observedInput) \defeq \lVert \mbz_m \rVert \lVert \observedInput \rVert \cdot \sigmoid\left( \frac{\mbz_m^\top \observedInput}{\lVert \mbz_m \rVert \lVert \observedInput \rVert } \right).
\end{equation}
Now, since this 
function
is itself zonal, 
it can be represented
in terms of the
spherical harmonics as in \cref{eq:mercer}.
Let $\featureCoefficient_{\level}$ denote its 
associated Fourier coefficient.
Thus,
the inter-domain 
features can be defined as $\phi_{m} \defeq \hiddenLayer_{m}$,
which
leads to the covariance
\begin{equation} \label{eq:Kuu-activation-features}
  \left[\Kuu\right]_{mm'} = \sum_{\substack{\level=0: \\ \kernelCoefficient_\level \neq 0}}^{\infty} \frac{\featureCoefficient_\level^2}{\kernelCoefficient_\level} \frac{\level + \alpha}{\alpha} C_{\level}^{(\alpha)} \left ( \frac{\mbz_{m}^\top \mbz_{m'}}{\lVert \mbz_{m} \rVert \lVert \mbz_{m'} \rVert} \right ),
\end{equation}
where $\kernelCoefficient_\level$ denotes the Fourier coefficients associated with kernel $k$.
This formulation, which we refer to as \textsc{activated} \gls{SVGP},
has been shown to produce
competitive 
results, especially when
multiple \textsc{activated} \gls{SVGP} layers are stacked to 
form a \gls{DGP}~\citep{damianou2013deep}, 
such that the propagation of the predictive means
closely 
emulates a forward-pass through a deep \gls{NN}.

Despite these favorable properties, \textsc{activated} \glspl{SVGP} have several limitations when it comes to 
their use
with common covariance functions. 
Before 
elaborating on them
in \cref{sec:methodology}, we discuss the orthogonally-decoupled \gls{GP} framework
on which our proposed extension
relies.

\begin{figure}[t]
  \begin{center}
  \centerline{\includegraphics[width=\linewidth]{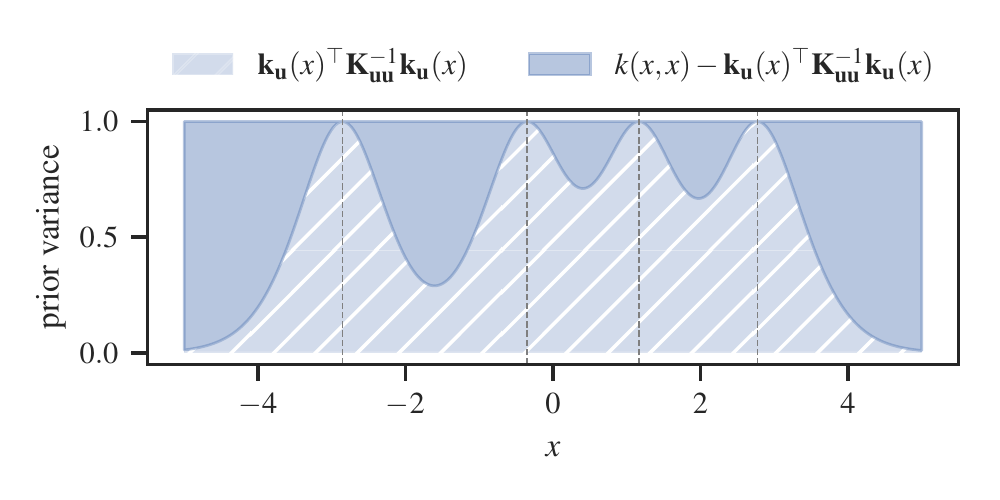}}
  \vskip -0.2in
  \caption{Prior variance deconstructed. 
    The prior variance 
    of $f(\observedInput)$ 
    is $\kernel(\observedInput, \observedInput) = \alpha$ for kernel amplitude $\alpha = 1$, 
    which can be 
    decomposed as the sum of the prior variances of $\parallelFunc(\observedInput)$ 
    and $\perpFunc(\observedInput)$.
    Vertical dashed lines indicate the 
    location of inducing inputs $\mbz_m$ for $m = 1, \dotsc, 4$.
    At these locations, the variance of $\parallelFunc(\observedInput)$ is 1 while 
    that
    of $\perpFunc(\observedInput)$ is 0.}
  \label{fig:variance-decomposition}
  \end{center}
  \vskip -0.3in
\end{figure}

\subsection{Orthogonally Decoupled Inducing Points}
\label{sub:orthogonally_decoupled_inducing_points}

Recent work has improved the efficiency of sparse \gls{GP} methods through the 
structured decoupling of inducing variables \citep{cheng2017variational,salimbeni2018orthogonally,shi2020sparse}. 
This not only enables the use of more variables at a reduced computational 
expense but also allows for more flexibility in modelling the predictive mean 
and covariance independently.
We focus on the general framework of \citet{shi2020sparse}
under which its predecessors 
can be subsumed.
\begin{table}[t]
  \caption{Summary of notation: relationships between input locations and output variables.}
  \label{tab:legend}
  \vskip 0.15in
  \begin{center}
    \begin{small}
      \begin{tabular}{@{}llll@{}}
      \toprule
                         & $\altInducingInputCollection$ & $\observedInputCollection$ & $\inducingInputCollection$ \\ 
      \midrule
      $\func(\cdot)$     & $\altInducingVariables$       & $\funcValues$              & $\inducingVariables$       \\ 
      $\perpFunc(\cdot)$ & $\perpAltInducingVariables$    & $\perpFuncValues$         & - \\ 
      \bottomrule
      \end{tabular}
    \end{small}
  \end{center}
  \vskip -0.1in
\end{table}
In particular, let the random function $f(\observedInput)$ 
from~\cref{eq:gaussian-process-prior} be decomposed into the sum of 
two independent \glspl{GP}: $f(\observedInput) = \parallelFunc(\observedInput) + \perpFunc(\observedInput)$,
where
\begin{equation*}
  \parallelFunc(\observedInput) \sim \GP(0, \mbk_\inducingVariables^\top(\observedInput) \Kuu^{-1} \mbk_\inducingVariables(\observedInput')),  
  \perpFunc(\observedInput) \sim \GP(0, s(\observedInput, \observedInput'))
\end{equation*}
and let the covariance function $\schurKernel(\observedInput, \observedInput')$ 
be defined 
according to
the Schur complement of $\Kuu$,
\begin{equation*}
  \schurKernel(\observedInput, \observedInput') \defeq \kernel(\observedInput, \observedInput') - \mbk_\inducingVariables^\top(\observedInput) \Kuu^{-1} \mbk_\inducingVariables(\observedInput'),
\end{equation*}
where $\mbk_\inducingVariables$ is defined in \cref{eq:cross-covariance-vector}.
Intuitively, one can view $g$ as the projection of $f$ onto $\inducingVariables$, 
and $\perpFunc \perp \parallelFunc$, i.e. $\perpFunc$ is 
\emph{orthogonal}
to $\parallelFunc$~\citep{JMLR:v18:16-579} 
in the statistical sense of linear independence~\citep{rodgers1984linearly}. 
See \cref{fig:variance-decomposition} for an illustration of the priors 
of $\parallelFunc(\observedInput)$ and $\perpFunc(\observedInput)$.
Let $\perpFuncValues$ be the values of $\perpFunc$ at observed inputs $\observedInputCollection$, 
i.e. $\perpFuncValues \defeq \perpFunc(\observedInputCollection)$.
Then we have $p(\perpFuncValues) = \Normal(\mbzero, \Cff)$,
where $\Cff \defeq \Kff - \Qff$.
This allows one to reparameterize $\funcValues \sim p(\funcValues \g \inducingVariables)$ 
from~\cref{eq:prior-conditional}, for a given $\inducingVariables$,
as
\begin{equation} \label{eq:relationship-f-h}
  \funcValues = \Qfu \inducingVariables + \perpFuncValues,
  \qquad
  \perpFuncValues \sim p(\perpFuncValues).
\end{equation}
The model's joint distribution can now be written 
as $p(\mby, \perpFuncValues, \inducingVariables) = p(\mby \g \perpFuncValues, \inducingVariables) p(\perpFuncValues) p(\inducingVariables)$, 
where
$p(\mby \g \perpFuncValues, \inducingVariables) = \Normal(\mby \g \Qfu \inducingVariables + \perpFuncValues, \beta^{-1} \mbI)$.
Next, \emph{orthogonal} inducing variables $\altInducingVariables$, 
which represent the
values of $f$ at a collection of $K$ orthogonal inducing 
locations $\altInducingInputCollection \defeq 
[\altInducingInput_1 \cdots \altInducingInput_K]^\top$, are introduced.
Similarly, inducing variables $\perpAltInducingVariables$ 
represent
the values of $h$ at $\altInducingInputCollection$.
The reader may find it helpful to refer to \cref{tab:legend} for a summary of 
the relationships between the input locations and the output variables 
defined thus far.

\glsunset{ODVGP}

Now, by definition, 
$\altInducingVariables = \Qvu \inducingVariables + \perpAltInducingVariables$, 
where $\Qvu \defeq \Kvu \Kuu^{-1}$ which is analogous to the relationship 
between $\funcValues$ and $\perpFuncValues$ in \cref{eq:relationship-f-h}.
Therefore, 
one need only be concerned with the
treatment 
of $\perpAltInducingVariables$.
The joint distribution of the model augmented by the variables $\perpAltInducingVariables$ now 
becomes $p(\mby, \mbh, \inducingVariables, \perpAltInducingVariables) = p(\mby \g \mbh, \inducingVariables) p(\inducingVariables) p(\mbh, \perpAltInducingVariables)$,
where $p(\mbh, \perpAltInducingVariables) = p(\mbh \g \perpAltInducingVariables) p(\perpAltInducingVariables)$
for $p(\perpAltInducingVariables) = \Normal(\mbzero, \Cvv)$ 
and $p(\mbh \g \perpAltInducingVariables) = \Normal(\mbh \g \Cfv \Cvv^{-1} \perpAltInducingVariables, \Cff - \Cfv \Cvv^{-1} \Cvf)$,
with
\begin{align}
  \Cvf \defeq \Kvf - \Qvf, \quad
  \Qvf \defeq \Qvu \Kuu \Quf, \\
  \Cvv \defeq \Kvv - \Qvv, \quad
  \Qvv \defeq \Qvu \Kuu \Quv.
\end{align}
Let the variational distribution 
now be 
$q(\mbh, \inducingVariables, \perpAltInducingVariables) = p(\mbh \g \perpAltInducingVariables) q(\inducingVariables, \perpAltInducingVariables)$, 
where $q(\inducingVariables, \perpAltInducingVariables) \defeq q(\inducingVariables) q(\perpAltInducingVariables)$ 
and $q(\perpAltInducingVariables) \defeq \cN\left(\variationalLoc_{\altInducingVariables}, \variationalCov_{\altInducingVariables}\right)$  
for variational parameters $\variationalLoc_{\altInducingVariables} \in \bbR^K$ 
and $\variationalCov_{\altInducingVariables} \in \bbR^{K\times K}$ s.t. $\variationalCov_{\altInducingVariables} \succeq 0$. 
This gives the posterior 
predictive density $q(\funcValues_\ast) = \Normal(\mbmu_\ast, \mbSigma_{\ast\ast})$, 
where
\begin{align}
  \mbmu_\ast 
  & \defeq \Q{\ast}{\inducingVariables} \variationalLoc_{\inducingVariables}  + \C{\ast}{\altInducingVariables} \Cvv^{-1} \variationalLoc_\altInducingVariables, \label{eq:solvegp-predictive-mean} \\
  \begin{split} \label{eq:solvegp-predictive-covariance}
    \mbSigma_{\ast\ast} 
    & \defeq \K{\ast}{\ast} + \Q{\ast}{\inducingVariables} (\variationalCov_{\inducingVariables} - \Kuu) \Q{\inducingVariables}{\ast} \\
    & \qquad + \C{\ast}{\altInducingVariables} \Cvv^{-1} (\variationalCov_{\altInducingVariables} - \Cvv) \Cvv^{-1} \C{\altInducingVariables}{\ast}.
  \end{split}
\end{align}
Thus seen, prediction incurs a cost of $\cO(M^3 + K^3)$ in this framework.
Like the 
so-called
\gls{ODVGP} framework of \citet{salimbeni2018orthogonally}, 
when seen from the dual \gls{RKHS} perspective, 
the predictive mean can be decomposed into a component that shares the same 
standard basis as the covariance, 
in addition to
another component that is \emph{orthogonal} to the 
standard basis.
However, this framework extends \gls{ODVGP} further by also
decomposing the predictive covariance into
parts corresponding to the standard and orthogonal bases.
Accordingly, setting $\variationalCov_{\altInducingVariables} = \Cvv$ 
recovers the \gls{ODVGP} framework, 
and further setting $\variationalLoc_\altInducingVariables = \mbzero$ recovers the 
standard \gls{SVGP} framework. 

\section{Methodology}
\label{sec:methodology}


\glsunset{RKHS}
\glsunset{RELU}


We begin this section by outlining some 
of the 
limitations of 
\textsc{activated} \glspl{SVGP} that preclude the use of numerous 
kernels and inducing features, 
not the least of which being popular choices of kernels such as the \gls{SE} 
kernel and the \Matern{} family of kernels, combined with \gls{NN} 
inducing features with \gls{RELU} activations.

\begin{figure}[t]
  \begin{center}
  \centerline{\includegraphics[width=\linewidth]{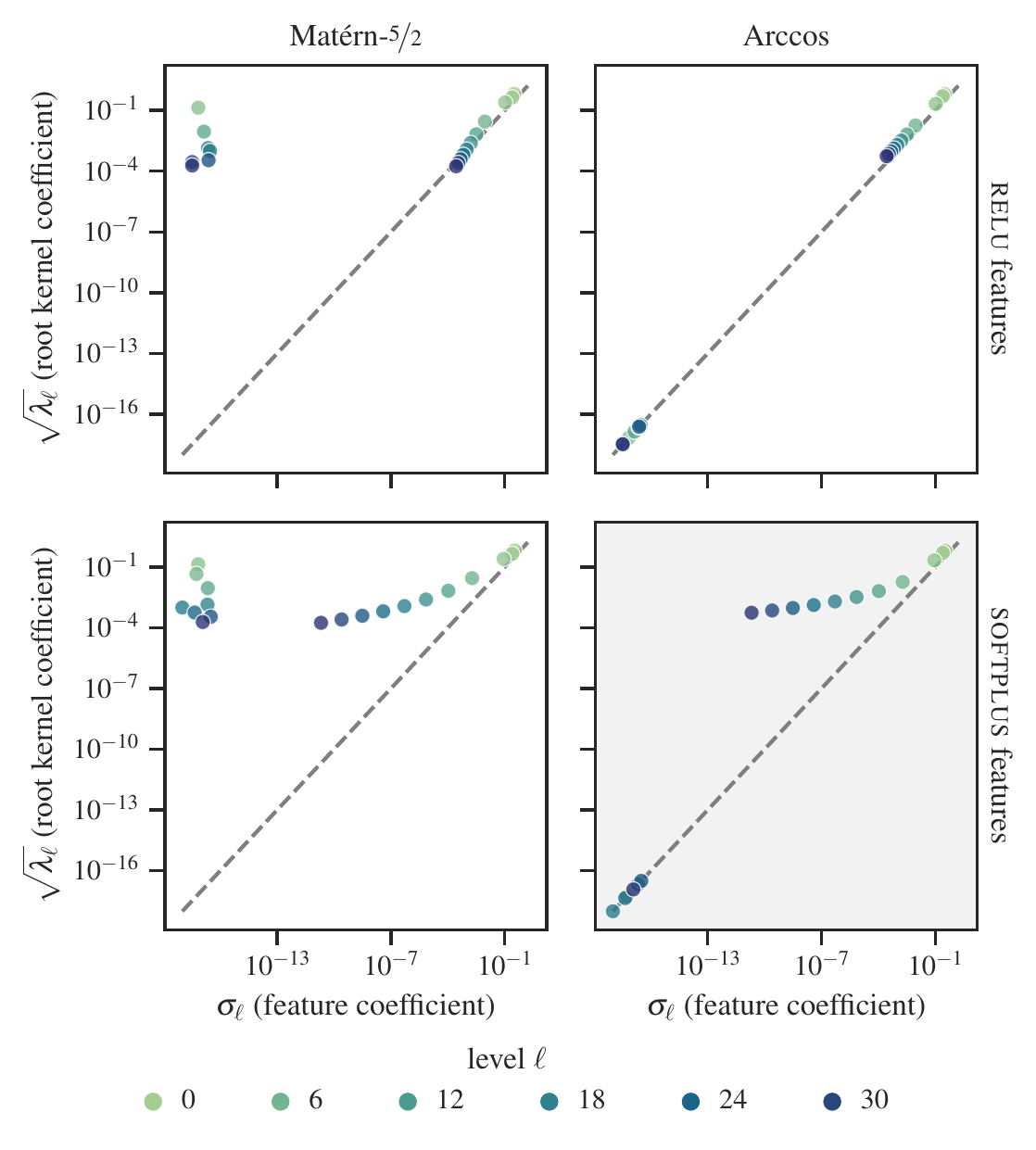}}
    \caption{Comparison of the Fourier coefficients of various kernels and activation features for increasing
      levels $\level = 1, \dotsc, 35$.}
    \label{fig:fourier-coefficients}
  \end{center}
  \vskip -0.3in
\end{figure}

\begin{figure}[t]
  \begin{center}
  \centerline{\includegraphics[width=\linewidth]{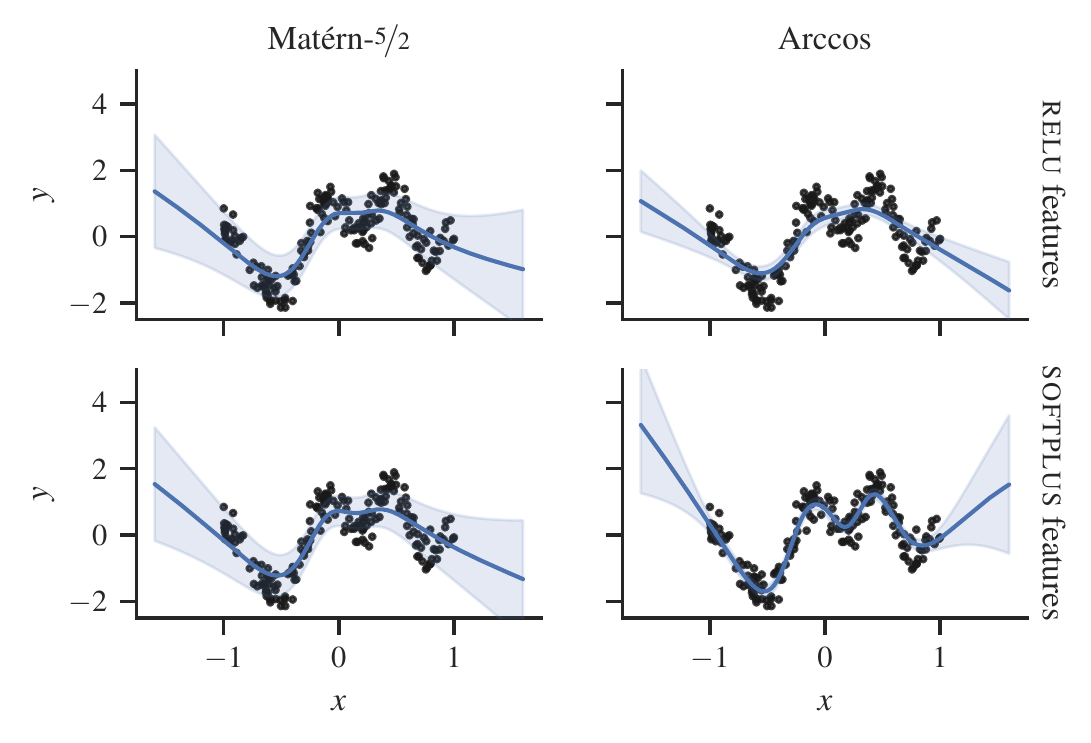}}
  \vskip -0.1in
  \caption{Posterior predictives of \textsc{activated} \gls{SVGP} models various kernels and activation features and $L=16$ levels.}
  \label{fig:predictives-levels-16}
  \end{center}
  \vskip -0.3in
\end{figure}

The root cause of these issues can be seen in~\cref{fig:fourier-coefficients}, where the 
Fourier coefficients of various combinations of kernels and 
activation 
features
are visualized.
Specifically, for each combination,
we compare the 
(root of the) kernel coefficients $\sqrt{\kernelCoefficient_{\level}}$ 
against the feature coefficients $\featureCoefficient_{\level}$ at increasing
levels $\level = 1, \dotsc, 35$.
The posterior predictives that result from fitting \textsc{activated} \gls{SVGP} models 
with these combinations are shown in \cref{fig:predictives-levels-16}.
We consider the \Matern{-$\nicefrac{5}{2}$} kernel as our running example, 
but the analysis extends to all stationary kernels.

\todo{Update symbol for feature coefficients in figure labels.}

{\bfseries Spectra mismatch.}
For the \Matern{} kernel (left column of panes 
in~\cref{fig:fourier-coefficients,fig:predictives-levels-16}), 
we see that there are multiple levels $\level$
at which the feature coefficients are zero while the corresponding kernel 
coefficients are nonzero. 
Such discrepancies in the spectra 
yields a poor \Nystroem{} 
approximation $\Qff$ that fails to fully capture the prior 
covariance $\Kff$ induced by the kernel, which 
subsequently
leads to the 
overestimation of the predictive variance and therefore a suboptimal \gls{ELBO}. 
In contrast, the Arccos kernel does not
suffer from this pathology.

{\bfseries RKHS inner product.} 
The \gls{RKHS} inner product %
associated with 
zonal kernels
in general is
a series 
consisting 
of ratios of Fourier coefficients.
Since the \gls{RELU} feature coefficients (top row of panes 
in~\cref{fig:fourier-coefficients,fig:predictives-levels-16}) decay at the same rate as the square root 
of the kernel coefficients, this results in a divergent series which in turn 
renders the \gls{RKHS} inner product indeterminate.
In contrast, the feature coefficients of the comparatively smoother \gls{SOFTPLUS} 
activation (bottom row of panes in~\cref{fig:fourier-coefficients,fig:predictives-levels-16}) decay at a 
much faster rate, and thus yields a well-defined \gls{RKHS} inner product. 
For the reasons outlined above, the work of \citet{dutordoir2021deep} 
restricted its scope to the
use of the
Arccos kernel 
in conjunction with the \gls{SOFTPLUS} activation 
(pane highlighted in gray in~\cref{fig:fourier-coefficients}).

{\bfseries Truncation error.} 
Lastly, as expected, the truncation of the series 
in \cref{eq:Kuu-activation-features} at some finite number  $L$ of 
spherical harmonic levels often leads to overly smooth predictive response 
surfaces and overestimation of the variance.

\subsection*{Spherical Features for Orthogonally-Decoupled GPs}

We propose extending the orthogonally-decoupled \gls{GP} framework (\cref{sub:orthogonally_decoupled_inducing_points}) to use inter-domain inducing features.
Accordingly, let $u_m \defeq \langle f, \phi_m \rangle_\cH$ and
$v_k \defeq \langle f, \psi_k \rangle_\cH$ for some arbitrary choices of $\phi_m, \psi_k \in \cH$.
This generalizes the framework of \citet{shi2020sparse} since, 
by the reproducing property, setting $\phi_m: \observedInput \mapsto k(\mbz_m, \observedInput)$ and $\psi_k: \observedInput \mapsto k(\altInducingInput_k, \observedInput)$
leads to standard 
inducing points, $u_m = f(\mbz_m), v_k = f(\altInducingInput_k)$. 

In particular, we define $\phi_m \defeq H_m$,
the $m$-th unit of the spherical activation layer (\cref{eq:spherical-activation-feature}) 
described in \cref{sub:neural_network_inducing_features}, and 
$\psi_k(\observedInput) \defeq k(\altInducingInput_k, \observedInput)$.
The posterior predictive of the model described 
in \cref{sub:orthogonally_decoupled_inducing_points}, summarized
by \cref{eq:solvegp-predictive-mean,eq:solvegp-predictive-covariance},
is fully determined by the covariances $\Kff, \Kuf, \Kvf, \Kuu, \Kvu$ and $\Kvv$.
Recall that $\left[\Kuf\right]_{mn} = \left[\mbk_\inducingVariables(\observedInput_n)\right]_m$  
and $\Kuu$ is precisely as expressed in \cref{eq:Kuu-activation-features}.
We have
\begin{align*}
  \left[\Kvf\right]_{kn} & \defeq \Cov{v_{k}}{f(\observedInput_{n})} = \kernel(\mbw_{k}, \observedInput_{n}), \\
  \left[\Kvu\right]_{km} & \defeq \Cov{v_{k}}{u_{m}} = \phi_{m}(\mbw_{k}), \\
  \left[\Kvv\right]_{kk'} & \defeq \Cov{v_{k}}{v_{k'}} = \kernel(\mbw_{k}, \mbw_{k'}).
\end{align*}
Note that the cross-covariance $\Kvu$ between $\mbu$ and $\mbv$ can be 
interpreted as the forward-pass of the orthogonal 
pseudo-input $\altInducingInput_k$ through the \gls{NN} activation $H_m$.
Crucially, these terms constitute the
orthogonal basis and provide additional degrees of flexibility, 
through free parameters $\altInducingInputCollection$,
that can compensate for  
errors 
remaining
from the original basis---in both the predictive mean and variance.
Suffice it to say, this is not the only possible choice but is one that 
possesses a number of appealing properties. 
We compare against a few other possibilities which we enumerate in \cref{sec:combinations}.


As discussed in \cref{sub:orthogonally_decoupled_inducing_points}, the addition
of $K$ inducing variables incurs a cost of $\cO(M^3 + K^3)$. 
More precisely:
suppose the exact cost is $C \cdot (M^3 + K^3)$ operations 
for some constant $C$ w.r.t. $M, K$.
Further, suppose $K=B \cdot M$ for some $B > 0$. 
Then there are a total of $(B+1) \cdot M$ inducing variables (orthogonal or otherwise)
and the cost becomes $(B^3+1) C \cdot M^3$.
By comparison, incorporating 
the same number of 
inducing variables in \gls{SVGP} costs $(B+1)^3 C \cdot M^3$. 
That is, 
this approach leads to a $(B^3+1)$-fold increase in the constant rather 
than a $(B+1)^3$-fold increase.
Concretely,
this means that
doubling the number of inducing variables 
doubles the constant in this approach, 
but leads to an \emph{eight-fold} increase in \gls{SVGP}.
While such a difference vanishes asymptotically for large $M$ and $K$, 
it still has a considerable impact for modest sizes ($M,K < 1,000$) that 
are feasible
in practice.
Thus seen, incorporating an orthogonal basis spanned by $K$ inducing variables
is a more cost-effective strategy for improving \textsc{activated} \gls{SVGP} 
than increasing 
$M$ or the truncation level $L$. 
\todo{We never explicitly described the cost of increasing the truncation level $L$! It is expensive!}

\begin{figure*}[t]
  \centering
  \begin{subfigure}[t]{0.49\textwidth}
    \centering
    \includegraphics[width=\linewidth]{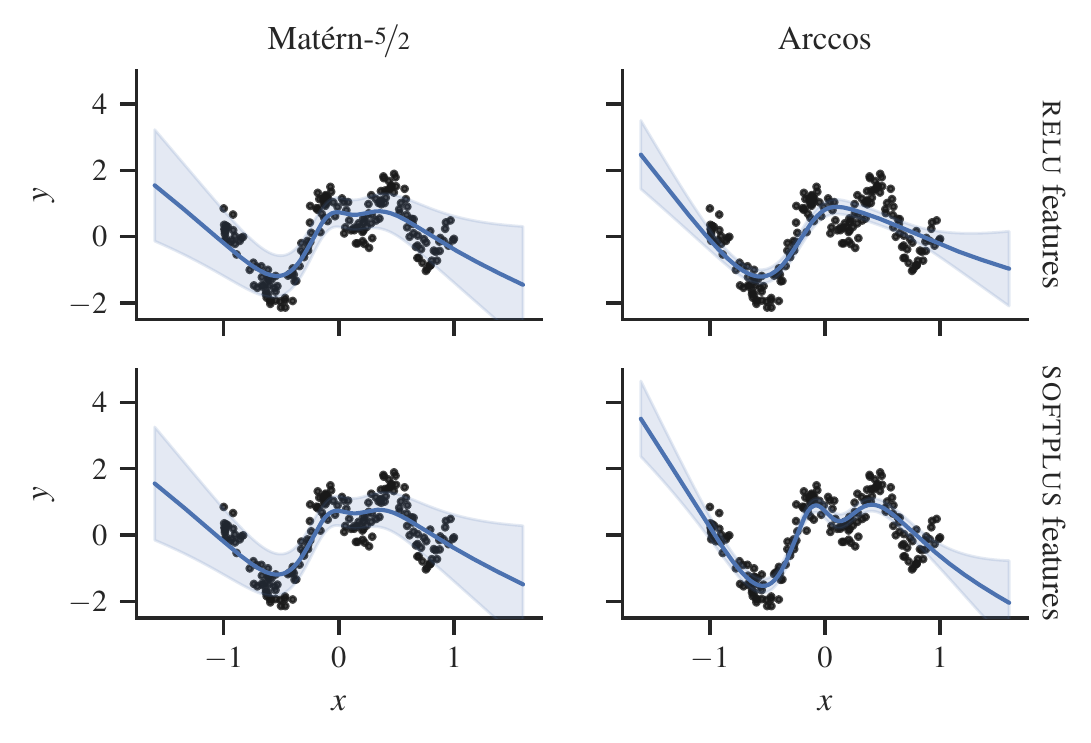}
    \caption{Inducing activation features with $L=8$ levels.}
    \label{fig:predictives-levels-8-svgp}
  \end{subfigure}
  \begin{subfigure}[t]{0.49\textwidth}
    \centering
    \includegraphics[width=\linewidth]{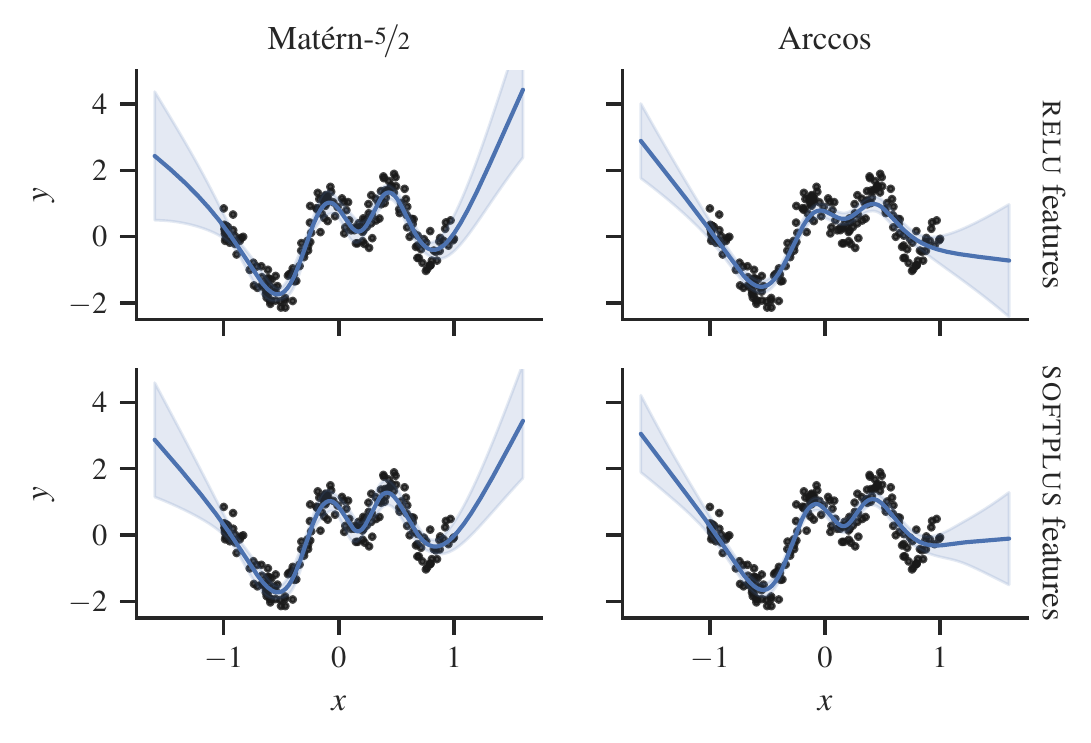}
    \caption{Inducing activation features with $L=8$ levels and $K=8$ orthogonal 
      bases (\emph{our method}).}
    \label{fig:predictives-levels-8-solvegp}
  \end{subfigure}
  \caption{
    Posterior predictives of \textsc{activated} \gls{SVGP}
    with various kernels and activation features 
    on the 1D Snelson dataset;
    \emph{black circular markers} represent the observations;
    \emph{blue solid lines} and \emph{shaded regions} denote the mean and 
    the $\pm2$ standard deviations, resp.}
  \label{fig:predictives-levels-8}
\end{figure*}

\begin{figure*}[t]
  \centering
  \begin{subfigure}[t]{0.49\textwidth}
    \centering
    \includegraphics[width=\linewidth]{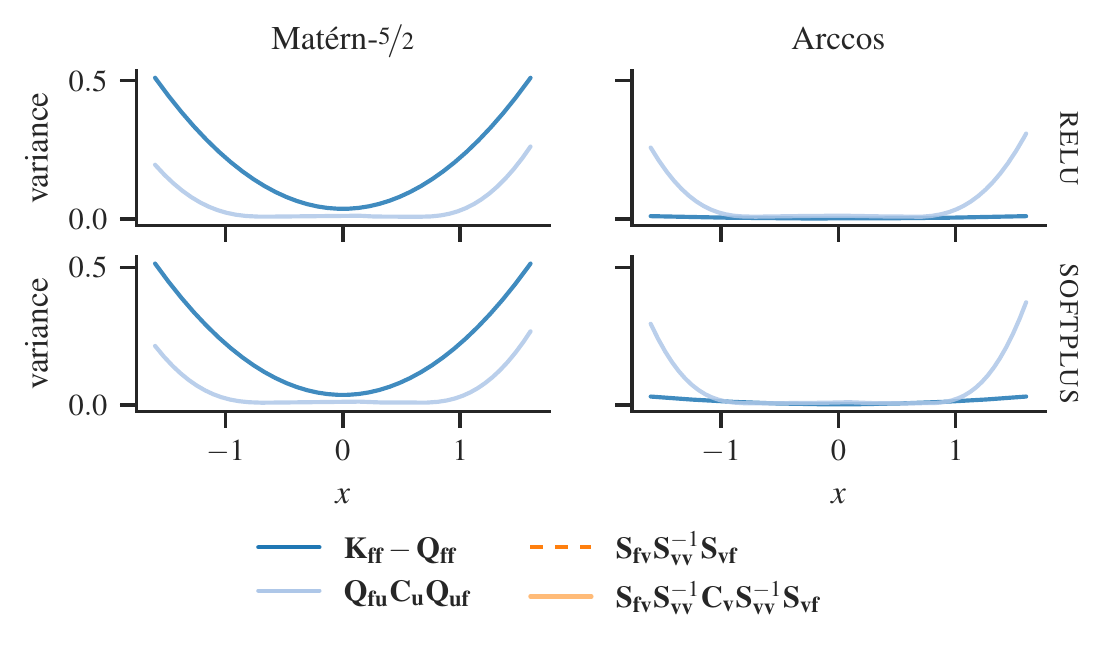}
    \caption{Inducing activation features with $L=8$ levels.}
    \label{fig:predictives-variances-levels-8-svgp}
  \end{subfigure}
  \begin{subfigure}[t]{0.49\textwidth}
    \centering
    \includegraphics[width=\linewidth]{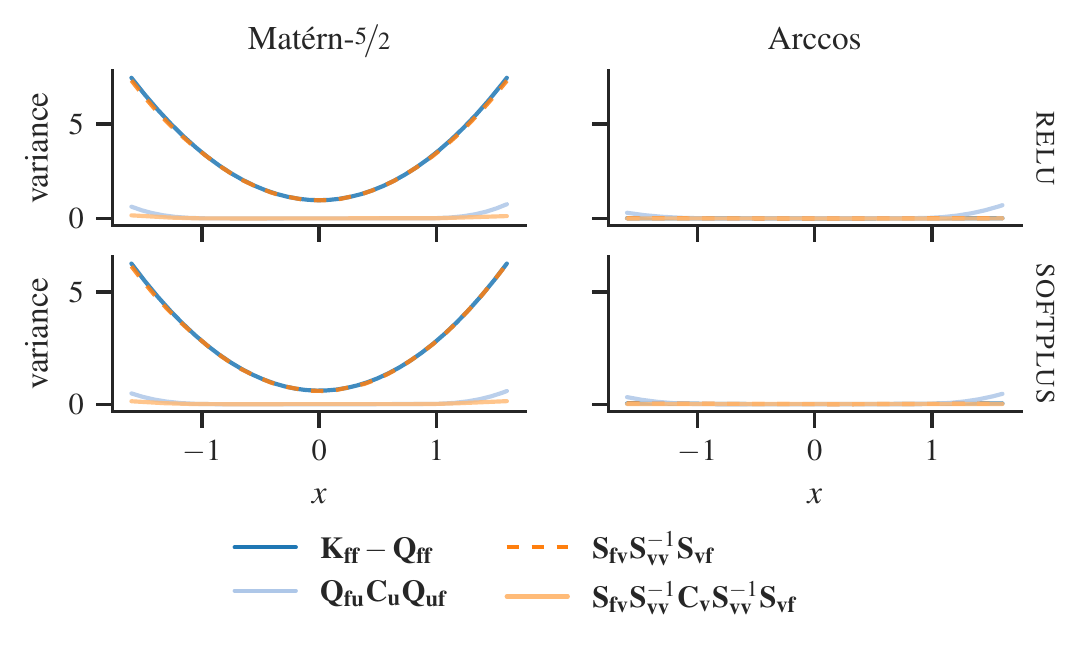}
    \caption{Inducing activation features with $L=8$ levels and $K=8$ orthogonal 
      bases (\emph{our method}).}    
    \label{fig:predictives-variances-levels-8-solvegp}
  \end{subfigure} 
  \caption{Decomposition of the posterior predictive variances
    of \gls{SVGP} 
    with various kernels and activation features 
    on the 1D \textsc{snelson} dataset (see~\cref{fig:predictives-levels-8})
    into its constituent terms;
    the additive terms that constitute the predictive variance are indicated 
    by \emph{solid} lines, while the subtractive terms 
    are indicated by \emph{dashed} lines;
    terms that constitute the predictive variance of the original \gls{SVGP} 
    model~\citep{titsias2009variational} have a \emph{blue} hue,
    while additional terms introduced by the orthogonally-decoupled 
    model~\citep{shi2020sparse} have an \emph{orange} hue.}
  \label{fig:predictive-variances}
\end{figure*}

\section{Experiments}
\label{sec:experiments}

\glsunset{UCI}

We describe the experiments conducted to empirically validate our approach.
The open-source implementation of our method can be found at:~\href{https://github.com/ltiao/spherical-orthogonal-gaussian-processes}{\sffamily ltiao/spherical-orthogonal-gaussian-processes}.
Further information concerning the experimental set-up and various 
implementation details can be found in \cref{sec:experimental_set_up_and_implementation_details}.

\subsection{Synthetic 1D Dataset}
\label{sub:synthetic_problems}


We 
highlight some notable properties of our method on the 
one-dimensional dataset of \citet{snelson2007local}.

First we fit \textsc{activated} \gls{SVGP} models 
with different combinations of kernels and 
activation
features 
using $L=8$ 
truncation
levels. 
The resulting posterior predictives
are shown in \cref{fig:predictives-levels-8}.
More specifically, in \cref{fig:predictives-levels-8-svgp}, 
we see that none of the model fits are particularly tight
due in part to truncation errors, since we are using relatively few levels.
This is especially true of the \Matern{} kernel (left column of panes), 
which results in a posterior that is not only 
too
smooth but also clearly suffering from an overestimation of 
the 
variance.
A conceptually straightforward way to improve performance is to increase the 
truncation level.
Accordingly, \cref{fig:predictives-levels-16} (introduced earlier in \cref{sec:methodology}) 
showed results from effectively the exact same set-up, but with twice the 
number of levels ($L=16$). 
With this increase, we see a clear improvement in the Arccos-\gls{SOFTPLUS} 
case,
but no discernible difference in the other combinations. 
Notably, the overestimation of the variances in the \Matern{} kernel 
persists.
By comparison, \cref{fig:predictives-levels-8-solvegp} shows results from 
using $L=8$ 
truncation
levels,
but with the addition 
of $K=8$ orthogonal inducing variables.
Remarkably, incorporating just a handful of these variables 
produces
substantial improvements, not least for the \Matern{} kernel.

\Cref{fig:predictive-variances} 
offers a deeper insight into the underlying mechanisms that contribute to these 
improvements.
Here we plot the predictive 
variance (\cref{eq:solvegp-predictive-covariance})
in terms of its constituent parts.
In \cref{fig:predictives-variances-levels-8-svgp}, we see that 
the variance estimate with \Matern{} kernels is heavily distorted by
large spurious contributions in the $\Kff - \Qff$ 
term (\emph{dark blue solid line}), which is caused by 
the pathology
described in \cref{sec:methodology}. 
On the other hand, in \cref{fig:predictives-variances-levels-8-solvegp}, 
such spurious contributions also 
appear,
but are offset by the subtractive term $\C{\funcValues}{\altInducingVariables} \Cvv^{-1} \C{\altInducingVariables}{\funcValues}$   
(\emph{dark orange dashed line}).
This term constitutes the orthogonal basis, and provides added flexibility 
that is 
effective at nullifying errors introduced by the original
basis.


\begin{figure}[t]
\centering
\includegraphics[width=\linewidth]{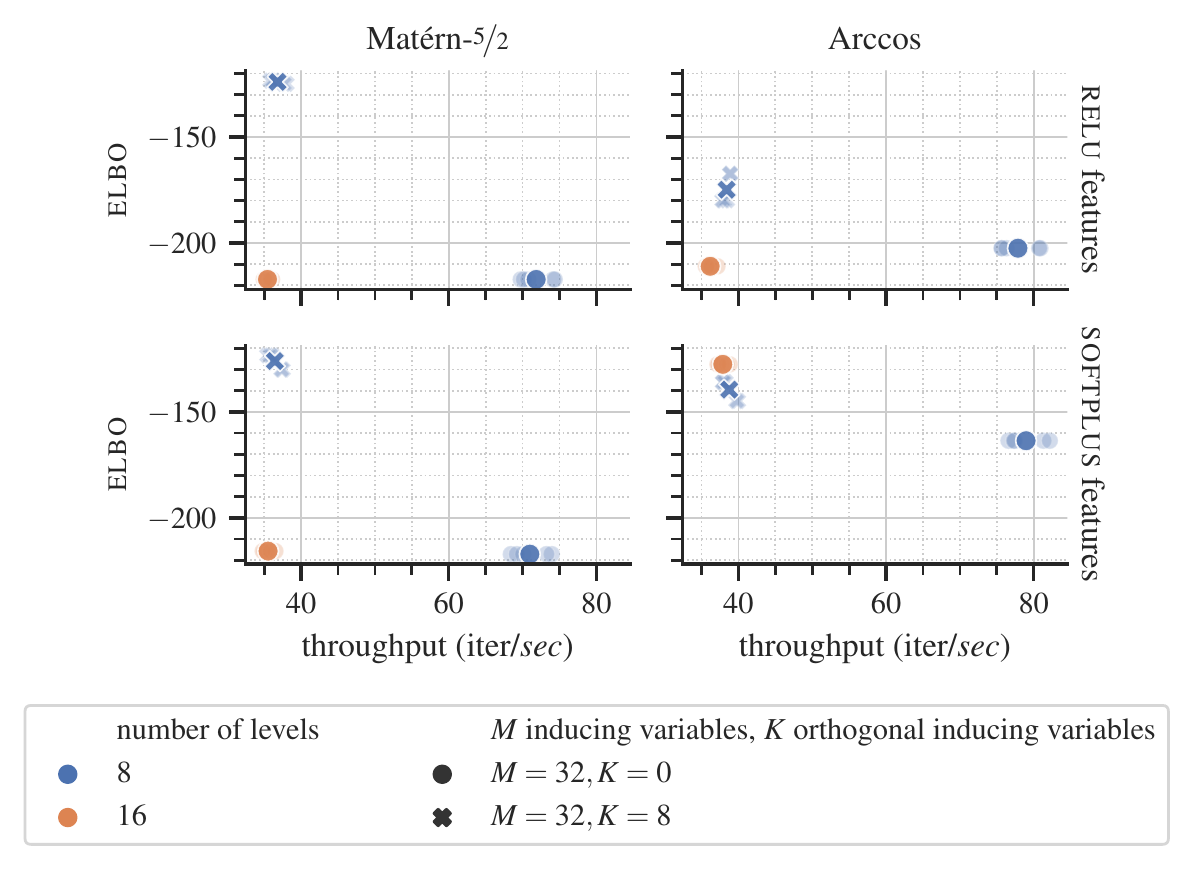}
\caption{The \gls{ELBO} and throughput (of model fitting) for various kernels 
  and activation features and the configurations visualized in 
  \cref{fig:predictives-levels-16}, \cref{fig:predictives-levels-8-svgp,fig:predictives-variances-levels-8-svgp},
  and \cref{fig:predictives-levels-8-solvegp,fig:predictives-variances-levels-8-solvegp}; 
  markers with low opacity represent the individual runs, 
  while markers with high opacity represent the mean of each group.}
\label{fig:snelson-elbo}
\vskip -0.2in
\end{figure}

Each of the three variations discussed above are repeated 5 times, 
and some quantitative results are summarized in \cref{fig:snelson-elbo}. 
Specifically, we report the \gls{ELBO} 
and the \emph{throughput}, i.e. the average number 
of optimization iterations completed per second.
The \textsc{activated} \gls{SVGP} with $L=8$ truncation levels, as seen 
in \cref{fig:predictives-levels-8-svgp,fig:predictives-variances-levels-8-svgp},
is represented by the \emph{blue circular markers}.
The model resulting from doubling the number of levels $L=16$, as seen 
in \cref{fig:predictives-levels-16}, is represented by the \emph{orange circular markers}. 
As discussed, this leads to an improvement in the Arccos-\gls{SOFTPLUS} case, 
but to modest or no improvements otherwise. 
However, we can now see that this has come at a significant computational 
expense, as the throughput has reduced by roughly half.
On the other hand, the model resulting from retaining the same truncation level
but incorporate an orthogonal basis consisting of $K=8$ variables,
as seen in \cref{fig:predictives-levels-8-solvegp,fig:predictives-variances-levels-8-solvegp},
is represnted by the \emph{blue cross markers}. 
This can be seen to have
roughly 
the same 
footprint
as doubling the truncation level, but leads to a considerably improved model fit, 
especially in cases involving the \Matern{} kernel (the only exception is in 
the Arccos-\gls{SOFTPLUS} case, where doubling the truncation level retains a 
slight advantage).
\todo{Add one short sentence to summarize the implications}
All told, 
incorporating an orthogonal basis has roughly the same cost as doubling the 
truncation level but leads to significantly better performance improvements.

\subsection{Regression on UCI Repository Datasets}
\label{sub:regression_on_uci_repository_datasets}

We evaluate our method on a number of well-studied regression problems from 
the \gls{UCI} repository of datasets~\cite{Dua:2019}.
In particular, we consider the \textsc{yacht}, \textsc{concrete}, \textsc{energy}, 
\textsc{kin8nm} and \textsc{power} datasets.
Additional results on the larger datasets from this collection can be found 
in \cref{sub:results_on_larger_uci_repository_datasets}.

We fit variations of \gls{SVGP} with the Arccos, \Matern{}, and \gls{SE} kernels, 
and (a) 
standard
inducing points, and inter-domain inducing features 
based on (b) \gls{RELU}- and (c) \gls{SOFTPLUS}-activated 
inducing features. 
For each of these 
variants, we consider three combinations 
of base and orthogonal inducing variables:
(\mplred{i}-\mplblue{ii}) 128 and 256 base inducing variables (and no orthogonal inducing variables), 
and (\mplgreen{iii}) 128 base inducing variables with 128 orthogonal inducing variables.
The 
activation features are 
truncated
at $L = 6$ levels.
Our proposed method is represented by the combinations 
consisting of \gls{RELU}- and \gls{SOFTPLUS}-activated 
features 
with orthogonal inducing variables (b-c,\mplgreen{iii}). 
The remaining combinations, against which we 
benchmark,
correspond to
the original \gls{SVGP} (a,\mplred{i}-\mplblue{ii}) \cite{titsias2009variational},
\gls{SOLVEGP} (a,\mplgreen{iii}) \cite{shi2020sparse}, and
\textsc{activated} \gls{SVGP} (b-c,\mplred{i}-\mplblue{ii}) \cite{dutordoir2021deep}.

\begin{figure*}[t]
  \vskip 0.2in
  \begin{center}
  \centerline{\includegraphics[width=\linewidth]{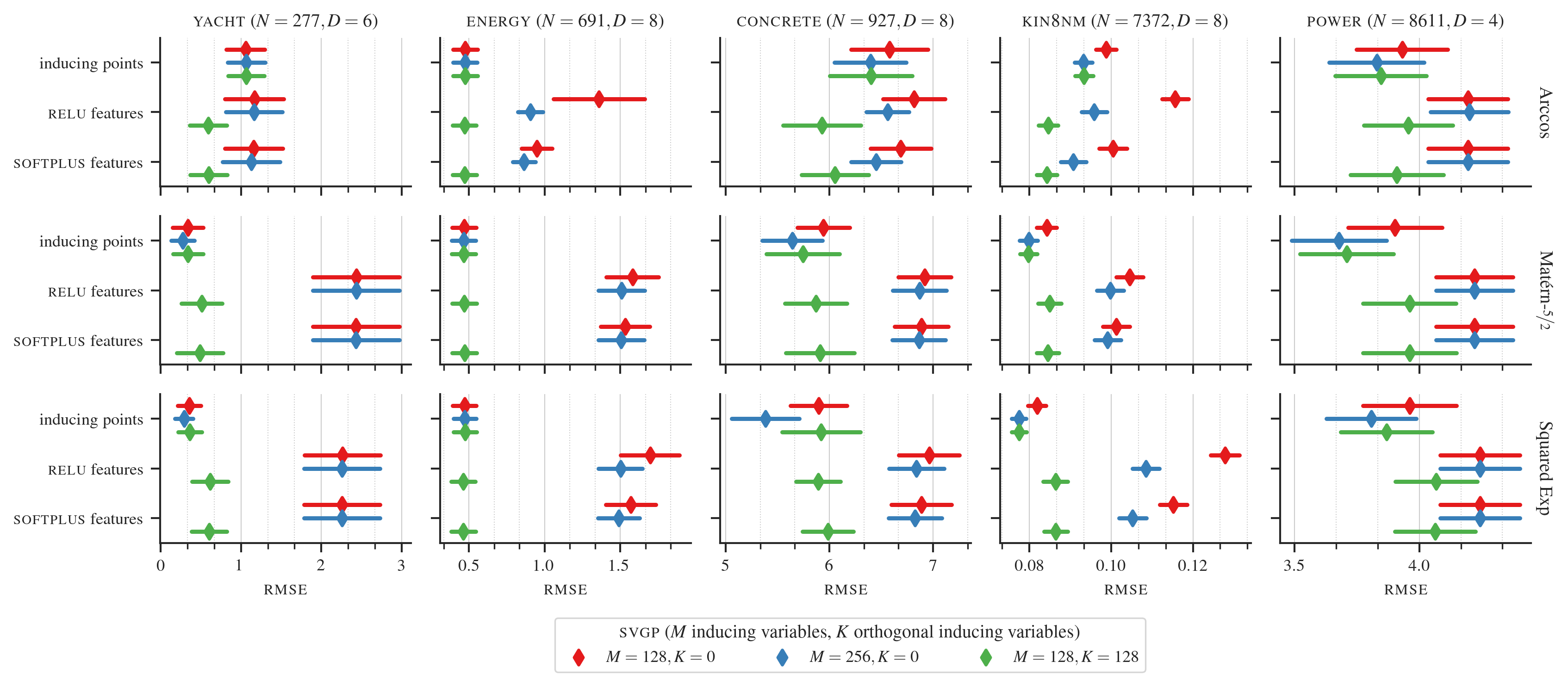}}
  \caption{Test \gls{RMSE} on regression problems from the \gls{UCI} repository of datasets for various kernels and activation features.
    Along the rows labeled \emph{``inducing points''}, 
    the red and blue markers ($\mplred{\blacklozenge}, \mplblue{\blacklozenge}$) 
    represent the original \gls{SVGP} model \cite{titsias2009variational},
    while the green markers ($\mplgreen{\blacklozenge}$) 
    represent \gls{SOLVEGP} \cite{shi2020sparse}.
    Along the remaining rows, 
    the red and blue markers ($\mplred{\blacklozenge}, \mplblue{\blacklozenge}$) 
    represent the \textsc{activated} \gls{SVGP} \cite{dutordoir2021deep},
    while the green markers ($\mplgreen{\blacklozenge}$) represent our 
    proposed approach.
  }
  \label{fig:uci-regression-rmse}
  \end{center}
  \vskip -0.2in
\end{figure*}

\begin{figure*}[t]
  \vskip 0.2in
  \begin{center}
  \centerline{\includegraphics[width=\linewidth]{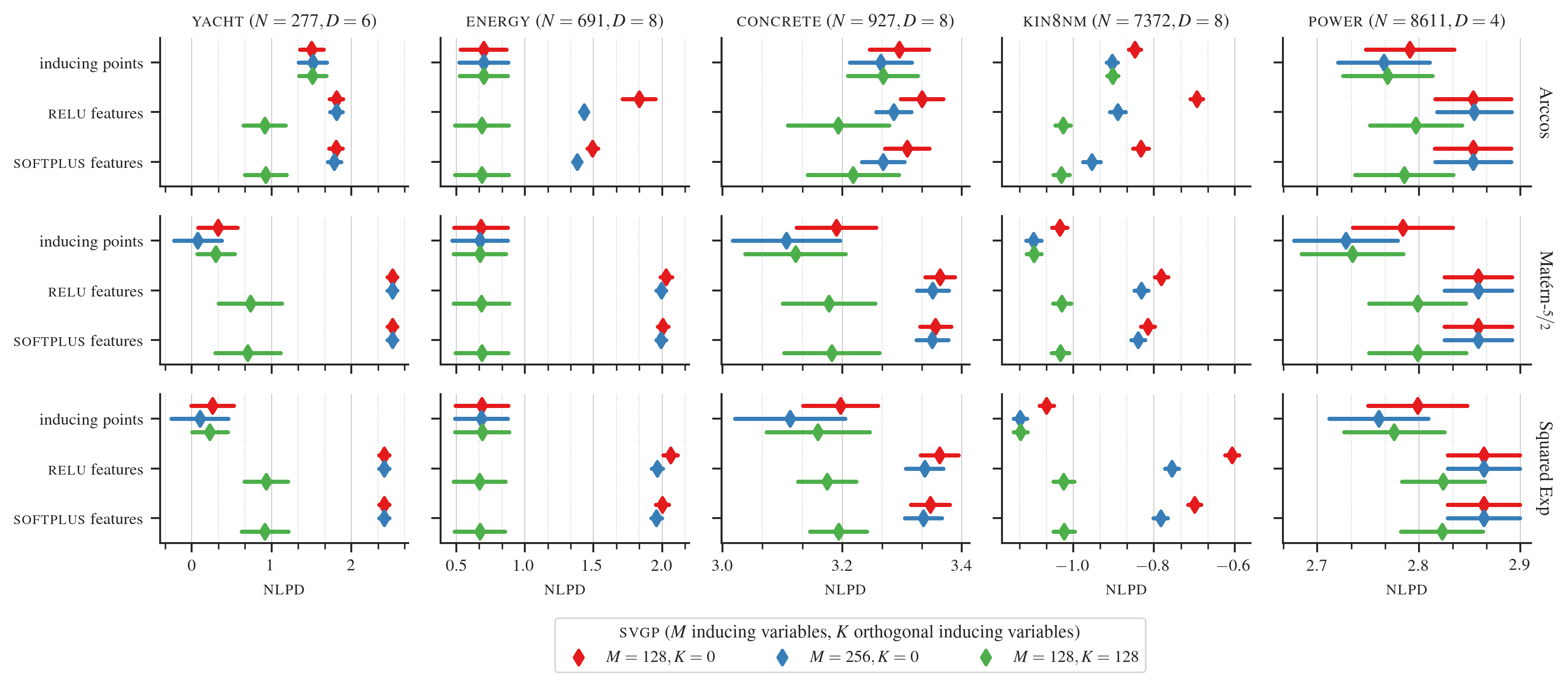}}
  \caption{Test \gls{NLPD} on regression problems from the \gls{UCI} repository of datasets for various kernels and activation features.
    Along the rows labeled \emph{``inducing points''}, 
    the red and blue markers ($\mplred{\blacklozenge}, \mplblue{\blacklozenge}$) 
    represent the original \gls{SVGP} model \cite{titsias2009variational},
    while the green markers ($\mplgreen{\blacklozenge}$) 
    represent \gls{SOLVEGP} \cite{shi2020sparse}.
    Along the remaining rows, 
    the red and blue markers ($\mplred{\blacklozenge}, \mplblue{\blacklozenge}$) 
    represent the \textsc{activated} \gls{SVGP} \cite{dutordoir2021deep},
    while the green markers ($\mplgreen{\blacklozenge}$) represent our 
    proposed approach.
  }
  \label{fig:uci-regression-nlpd}
  \end{center}
  \vskip -0.2in
\end{figure*}

To quantitatively assess performance, we report the 
test \gls{RMSE} and \gls{NLPD}, shown 
in \cref{fig:uci-regression-rmse,fig:uci-regression-nlpd}, respectively.
Unless otherwise stated, for each method and problem, we perform random 
sub-sampling validation by aggregating results from 5 repetitions
across 10\% held-out test sets.
Within the training set, the inputs and outputs are standardized, i.e. scaled 
to have zero mean and unit variance and subsequently restored to the original 
scale at test time.

\begin{figure}[t!]
  \centering
  \includegraphics[width=\linewidth]{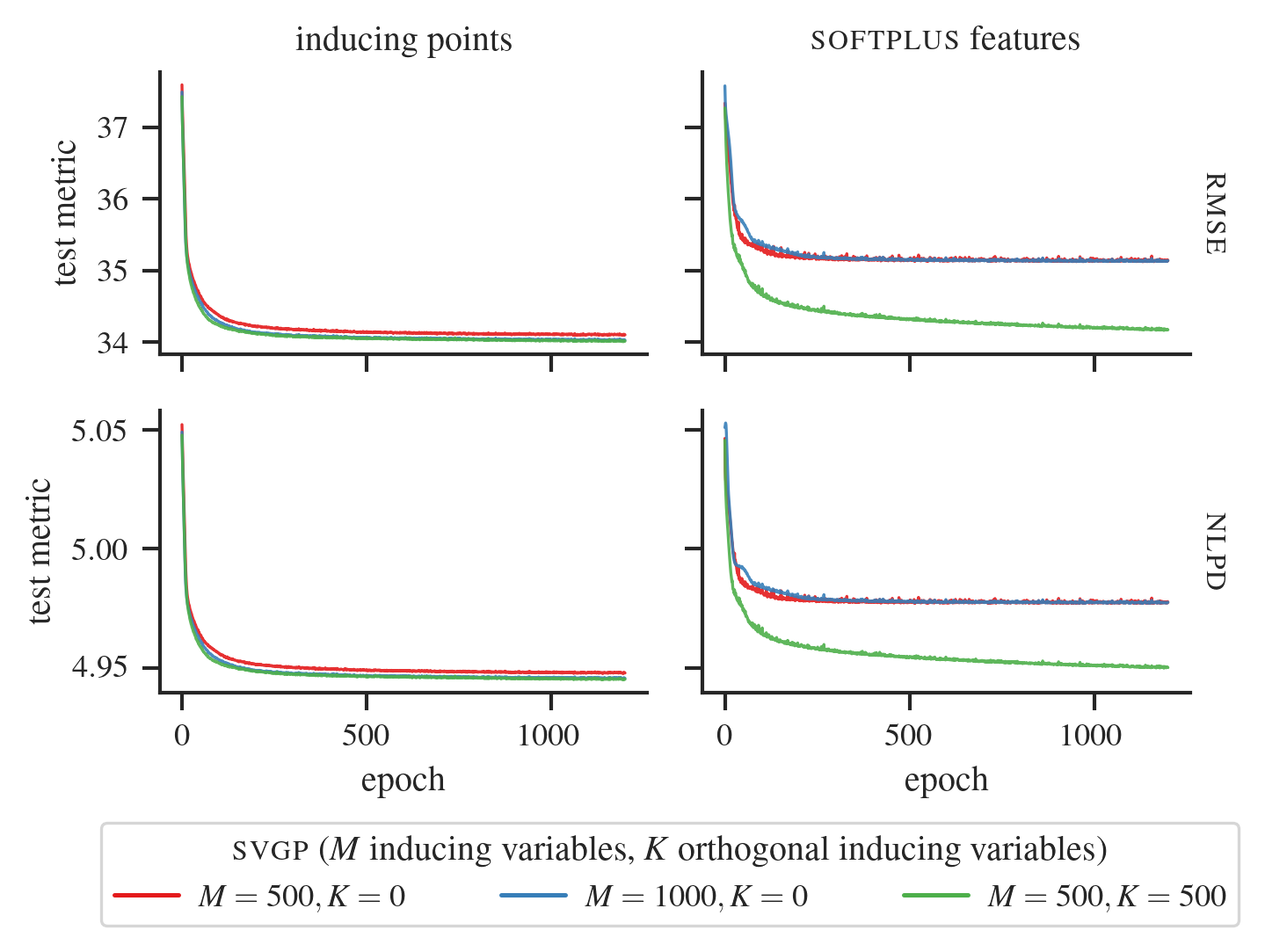}
  \caption{Test metrics, \gls{RMSE} and \gls{NLPD},
    on the large-scale 2008 U.S. airline delays dataset using 
    the \emph{Arccos} kernel with standard inducing points 
    and \gls{SOFTPLUS}-activated features.
    Along the column labeled ``inducing points'', 
    the red and blue lines (\mplred{---} and \mplblue{---}) 
    represent the mini-batch \gls{SVGP} \cite{hensman2013gaussian},
    while the green line (\mplgreen{---}) 
    represents \gls{SOLVEGP} \cite{shi2020sparse}.
    Along the column labeled ``\gls{SOFTPLUS} features'', 
    the red and blue lines (\mplred{---} and \mplblue{---}) 
    represent the \textsc{activated} \gls{SVGP} \cite{dutordoir2021deep},
    while the green line (\mplgreen{---}) represents our proposed approach.
  }
  \label{fig:airline-regression}
\vskip -0.2in
\end{figure}

We observe that, irrespective of the choice of kernel,
when using 
activation features, whether \gls{RELU}- or \gls{SOFTPLUS}-activated,
augmenting the model with  
orthogonal 
bases
significantly
improves performance, notably even more so
than doubling the number of 
base
inducing variables.
This can readily be seen across all datasets on both the \gls{NLPD} and \gls{RMSE} metrics. 
Further, with the Arccos kernel, it outperforms its counterparts based on 
standard inducing points across most datasets (the exception being 
the \textsc{power} dataset). 
With the \Matern{} and \gls{SE} kernels, it achieves results comparable to its 
standard inducing points counterparts in most datasets.

\subsection{Large-scale Regression on Airline Delays Dataset}
\label{sub:large_scale_regression_on_airline_delays_dataset}

Finally, we consider a large-scale regression dataset concerning U.S. commercial 
airline delays in 2008.
The task is to forecast the duration of delays in reaching the destination of a given flight, 
utilizing information such as the route distance, airtime, 
scheduled month, day of the week, and other relevant factors, as well as 
characteristics of the aircraft such as its age (number of years since deployment).
The complete dataset encompasses 5,929,413 flights, of which we 
randomly select 1M observations without replacement to form a subset that is 
more manageable but still considerable in scale.
Results on a reduced 100K subset can be found in \cref{sub:regression_on_airline_delays_dataset}.

To quantitatively assess performance, we report the 
test \gls{RMSE} and \gls{NLPD} evaluated on a $\nicefrac{1}{3}$ held-out test set.
The results are shown in the top and bottom rows of \cref{fig:airline-regression}, 
respectively.
Within the training set, the inputs and outputs are standardized, i.e. scaled 
to have zero mean and unit variance and subsequently restored to the original 
scale at test time.

Given the immense volume of data at hand, we are compelled to utilize mini-batch 
training for stochastic optimisation~\cite{hensman2013gaussian}.
To this end, we use the Adam optimizer~\cite{kingma2014adam} with its typical 
default settings (learning rate $\num{1e-3}$, $\beta_1=0.9, \beta_2=0.999$).
Our batch size is set to 5,000, and we train the models for a total of 1,200 epochs.

We fit variations of \gls{SVGP} with the Arccos kernel 
and (a) 
standard
inducing points
and (b) inter-domain inducing features based on \gls{SOFTPLUS}-activated
inducing features. 
For each of these 
variants, we consider three combinations 
of base and orthogonal inducing variables:
(\mplred{i}-\mplblue{ii}) 500 and 1,000 base inducing variables (and no orthogonal inducing variables), 
and (\mplgreen{iii}) 500 base inducing variables with 500 orthogonal inducing variables.
The 
activation features are 
truncated
at $L = 6$ levels.
Our proposed method is represented by the combination 
consisting of \gls{SOFTPLUS}-activated 
features 
with orthogonal inducing variables (b,\mplgreen{iii}). 
The remaining combinations, against which we 
benchmark,
correspond to
the mini-batch \gls{SVGP} (a,\mplred{i}-\mplblue{ii}) \cite{hensman2013gaussian},
\gls{SOLVEGP} (a,\mplgreen{iii}) \cite{shi2020sparse}, and
\textsc{activated} \gls{SVGP} (b,\mplred{i}-\mplblue{ii}) \cite{dutordoir2021deep}.

The outcomes are as expected when employing standard inducing points (left).
In particular, doubling the number of base inducing points from 500 to 1,000 
demonstrates significant improvements.
Furthermore, by using 500 base inducing points alongside 500 orthogonal 
inducing points, we 
achieve comparable performance to 
having 1,000 base inducing points, while enjoying improved computationally efficiency.
In contrast, 
when examinining the \textsc{activated} \gls{SVGP} model with \gls{SOFTPLUS} 
features (right), it's apparent that it underperforms compared to the 
original \gls{SVGP} counterparts. 
Furthermore, doubling the number of inducing features from 500 to 1,000 has 
virtually no effect.
However, by incorporating orthogonal bases into
the \textsc{activated} \gls{SVGP} model with 500 features 
following our proposed approach, we witness substantial improvements and 
achieve comparable performance to its standard inducing points counterparts.

\section{Conclusion}
\label{sec:conclusion}

We considered the use of inter-domain inducing features in the 
orthogonally-decoupled \gls{SVGP} framework, specifically, the 
spherical activation features,
and showed that this alleviates some of the practical issues and shortcomings 
associated with 
the \textsc{activated} \gls{SVGP} model.
We demonstrated the effectiveness of this approach by conducting empirical 
evaluations on several problems,
and showed that this leads to enhanced predictive performance over more 
computationally demanding alternatives such as increasing the truncation levels 
or the number of inducing variables.

Future work will explore alternative designs of inter-domain inducing features 
to construct new standard and orthogonal bases that provide additional 
complementary benefits.

\section*{Acknowledgements}

We are grateful to Thang Bui and Jiaxin Shi for
sharing their valuable insights and providing helpful discussions. 
We also extend our appreciation to the anonymous reviewers for their 
thoughtful feedback and constructive suggestions, which have significantly 
enhanced the quality 
our work.

\newpage

\bibliography{main,example_paper}
\bibliographystyle{icml2023}

\newpage
\appendix
\onecolumn

\section{Combinations}
\label{sec:combinations}

Recall that the prior over $\funcValues, \inducingVariables, \altInducingVariables$ is 
\begin{equation}
  \begin{bmatrix}
    \funcValues \\ \inducingVariables \\ \altInducingVariables
  \end{bmatrix}
  \sim
  \Normal \left ( 
  \begin{bmatrix}
    \mbzero \\ \mbzero \\ \mbzero
  \end{bmatrix}, 
  \begin{bmatrix}
    \Kff & \Kuf^\top & \Kvf^\top \\
    \Kuf & \Kuu      & \Kvu^\top \\
    \Kvf & \Kvu      & \Kvv 
  \end{bmatrix}
  \right )
\end{equation}

Other than the variational parameters, the ingredients necessary to compute the 
predictive for the model described in~\cref{sub:orthogonally_decoupled_inducing_points},
summarized by \cref{eq:solvegp-predictive-mean,eq:solvegp-predictive-covariance}, 
are
\begin{equation*}
  \Cff \defeq \Kff - \Qff, \qquad
  \Cvf \defeq \Kvf - \Qvf, \qquad
  \Cvv \defeq \Kvv - \Qvv,
\end{equation*}
where
\begin{equation*}
  \Qff \defeq \Qfu \Kuu \Quf, \qquad
  \Qvf \defeq \Qvu \Kuu \Quf, \qquad
  \Qvv \defeq \Qvu \Kuu \Quv,
\end{equation*}
and
\begin{equation*}
  \Qfu \defeq \Kfu \Kuu^{-1}, \qquad 
  \Qvu \defeq \Kvu \Kuu^{-1}.
\end{equation*}
In all the cases we shall discuss below, $\Kff$ will evaluate to
\begin{equation}
  \left[\Kff\right]_{nn'} 
  = \Cov{f(\observedInput_{n})}{f(\observedInput_{n'})}
  = \langle \kernel(\observedInput_{n}, \argdot), \kernel(\observedInput_{n'}, \argdot) \rangle_\cH 
  = \kernel(\observedInput_{n}, \observedInput_{n'}).
\end{equation}
Similarly, $\Kuf$ and $\Kvf$ will always simplify to
\begin{align}
  \left[\Kuf\right]_{mn}  
  & = \Cov{u_m}{f(\observedInput_n)}
    = \langle \phi_m, \kernel(\observedInput_{n}, \argdot) \rangle_\cH 
    = \phi_m(\observedInput_{n}) \label{eq:Kuf} \\
  \left[\Kvf\right]_{kn} 
  & = \Cov{v_k}{f(\observedInput_n)}
    = \langle \psi_k, \kernel(\observedInput_{n}, \argdot) \rangle_\cH  
    = \psi_k(\observedInput_n) \label{eq:Kvf}
\end{align}
The remaining covariances $\Kuu, \Kvu$ and $\Kvv$ reduce to different 
forms, depending on the choices of $\phi_m$ and $\psi_k$,
\begin{align}
  \left[\Kuu\right]_{mm'} & = \Cov{u_{m}}{u_{m'}} = \langle \phi_{m}, \phi_{m'} \rangle_\cH, \label{eq:Kuu}  \\
  \left[\Kvu\right]_{km}  & = \Cov{v_{k}}{u_{m}}  = \langle \psi_k, \phi_m \rangle_\cH, \\
  \left[\Kvv\right]_{kk'} & = \Cov{v_{k}}{v_{k'}} = \langle \psi_{k}, \psi_{k'} \rangle_\cH.
\end{align}

In general, we have
\begin{equation} \label{eq:Qvf-expanded}
  \left [ \Qvf \right ]_{kn} = \sum_{\substack{m=1, \\ m'=1}}^M \left [ \Kuu^{-1} \right ]_{mm'} \langle \psi_k, \phi_m \rangle_\cH \langle \phi_{m'},  k(\observedInput_n, \cdot) \rangle_\cH,
\end{equation}
and
\begin{equation}
  \left [ \Qvv \right ]_{kk'} = \sum_{\substack{m=1, \\ m'=1}}^M \left [ \Kuu^{-1} \right ]_{mm'} \langle \psi_k, \phi_m \rangle_\cH \langle \phi_{m'},  \psi_{k'} \rangle_\cH
\end{equation}

We enumerate the combinations in turn and discuss their benefits and drawbacks.

\todo{make a table}

\subsection{Base Inducing Variables as Standard Inducing Points} 
\label{sub:standard_inducing_points}

In this case, the base inducing variables are simply 
standard inducing points $u_m = f(\mbz_m)$, 
which is equivalent to 
defining $\phi_m: \observedInput \mapsto k(\mbz_m, \observedInput)$.
Therefore, we have
\begin{align}
  \left[\Kuf\right]_{mn}  
  & = \phi_m(\observedInput_{n})
    = k(\mbz_{m}, \observedInput_{n}), \\
  \left[\Kuu\right]_{mm'}
  & = \langle \phi_{m}, \phi_{m'} \rangle_\cH
    = \langle \kernel(\mbz_{m}, \argdot), \kernel(\mbz_{m'}, \argdot) \rangle_\cH 
    = k(\mbz_{m}, \mbz_{m'}).
\end{align}

\subsubsection{Orthogonal Inducing Variables as Standard Inducing Points}
  
In this case, the orthogonal inducing variables are simply  
standard inducing points $v_k = f(\altInducingInput_m)$, which is equivalent to 
defining $\psi_k: \observedInput \mapsto k(\altInducingInput_k, \observedInput)$.
Note that this combination is precisely the original method of \citet{shi2020sparse}.
Hence, we have
\begin{align}
  \left[\Kvf\right]_{kn} 
  & = \psi_k(\observedInput_n) 
    = \kernel(\altInducingInput_{k}, \observedInput_{n}), \\
  \left[\Kvu\right]_{km}
  & = \langle \psi_k, \phi_m \rangle_\cH 
    = \langle \kernel(\altInducingInput_k, \argdot), \kernel(\mbz_m, \argdot) \rangle_\cH
    = \kernel(\altInducingInput_k, \mbz_m), \\
  \left[\Kvv\right]_{kk'} 
  & = \langle \psi_{k}, \psi_{k'} \rangle_\cH 
    = \langle \kernel(\altInducingInput_{k}, \argdot), \kernel(\altInducingInput_{k'}, \argdot)) \rangle_\cH 
    = \kernel(\altInducingInput_{k}, \altInducingInput_{k'}).
\end{align}
This leads to
\begin{equation}
  \left [ \Qvf \right ]_{kn} = \sum_{m=1}^M \sum_{m'=1}^M \left [ \Kuu^{-1} \right ]_{mm'} k(\altInducingInput_k, \mbz_m) k(\mbz_{m'}, \observedInput_n),
\end{equation}
and $\Cvf$ does not simplify further.

\subsubsection{Orthogonal Inducing Variables as Spherical Harmonics}

Here, orthogonal inducing variables are defined through inter-domain features 
$v_k = \langle f, \psi_k \rangle_\cH$ where $\psi_k \defeq \sphericalHarmonic{\level}{j}$, 
the spherical harmonic of level $\level$ and order $j$, 
and $k$ indexes the pairs $(\level, j)$.
Hence, $\Kvf$ is as expressed in \cref{eq:Kvf},
\begin{align}
  \left[\Kvu\right]_{km} & = \langle \psi_k, \phi_m \rangle_\cH 
    = \langle \psi_k, \kernel(\mbz_m, \argdot) \rangle_\cH = \psi_k(\mbz_m) \label{eq:Kvu-u-standard-inducing-points-v-spherical-harmonics} \\
  \left[\Kvv\right]_{kk'} & = \langle \psi_{k}, \psi_{k'} \rangle_\cH 
    = \kernelCoefficient_k^{-1} \delta_{kk'}
\end{align}
This leads to
\begin{equation}
  \left [ \Qvf \right ]_{kn} = \sum_{m=1}^M \sum_{m'=1}^M \left [ \Kuu^{-1} \right ]_{mm'} \psi_k(\mbz_m) k(\mbz_{m'}, \observedInput_n)    
\end{equation}
and $\Cvf$ does not simplify further.

\subsubsection{Orthogonal Inducing Variables as Spherical {NN} Activations} 

Here, orthogonal inducing variables are defined through inter-domain features 
$v_k = \langle f, \psi_k \rangle_\cH$ 
where $\psi_k = H_k$, the $k$-th unit of a 
spherical \gls{NN} activation layer (\cref{eq:spherical-activation-feature}).
Again, $\Kvf$ is as expressed in \cref{eq:Kvf}, and we still 
get $\left[\Kvu\right]_{km} = \psi_k(\mbz_m)$ as 
in \cref{eq:Kvu-u-standard-inducing-points-v-spherical-harmonics} 
from the preceding case, 
but we now have
\begin{equation}
  \left[\Kvv\right]_{kk'} = \langle \psi_{k}, \psi_{k'} \rangle_\cH = \sum_{\substack{\level=0: \\ \kernelCoefficient_\level \neq 0}}^{\infty} \frac{\featureCoefficient_\level^2}{\kernelCoefficient_\level} \frac{\level + \alpha}{\alpha} C_{\level}^{(\alpha)} \left ( \frac{\altInducingInput_{k}^\top \altInducingInput_{k'}}{\lVert \altInducingInput_{k} \rVert \lVert \altInducingInput_{k'} \rVert} \right ),
\end{equation}
which leads to
\begin{equation}
  \left [ \Qvf \right ]_{kn} = \sum_{m=1}^M \sum_{m'=1}^M \left [ \Kuu^{-1} \right ]_{mm'} \psi_k(\mbz_m) k(\mbz_{m'}, \observedInput_n).
\end{equation}
and $\Cvf$ does not simplify further.

\subsection{Base Inducing Variables as Spherical Harmonics} 

Here, the base inducing variables are defined through inter-domain features 
$u_m = \langle f, \phi_m \rangle_\cH$ where $\phi_m \defeq \sphericalHarmonic{\level}{j}$, 
the spherical harmonic of level $\level$ and order $j$, and $m$ indexes the pairs $(\level, j)$.
In this case, $\left[\Kuf\right]_{mn} = \phi_m(\observedInput_{n})$ as in \cref{eq:Kuf}, and
\begin{equation}
  \left[\Kuu\right]_{mm'} 
    = \langle \phi_{m}, \phi_{m'} \rangle_\cH
    = \kernelCoefficient_{m}^{-1} \delta_{mm'}.
\end{equation}
Therefore, we can already simplify $\Qvf$ from \cref{eq:Qvf-expanded} as
\begin{equation} \label{eq:Qvf-u-spherical-harmonics}
  \left [ \Qvf \right ]_{kn} 
  = \sum_{m=1}^M \sum_{m'=1}^M  \delta_{mm'} \kernelCoefficient_m \langle \psi_k, \phi_m \rangle_\cH \phi_{m'}(\observedInput_n) 
  = \sum_{m=1}^M \kernelCoefficient_m \langle \psi_k, \phi_m \rangle_\cH \phi_{m}(\observedInput_n)
\end{equation}

\subsubsection{Orthogonal Inducing Variables as Standard Inducing Points} 
\label{ssub:u-spherical-harmonics-v-standard-inducing-points}

In this case, the orthogonal inducing variables are simply  
standard inducing points $v_k = f(\altInducingInput_m)$, which is equivalent to 
defining $\psi_k: \observedInput \mapsto k(\altInducingInput_k, \observedInput)$.
Hence, we have
\begin{align}
  \left[\Kvf\right]_{kn} 
  & = \psi_k(\observedInput_n) 
    = \kernel(\altInducingInput_{k}, \observedInput_{n}) \\
  \left[\Kvu\right]_{km}
  & = \langle \psi_k, \phi_m \rangle_\cH 
    = \langle \kernel(\altInducingInput_k, \argdot), \phi_m \rangle_\cH
    = \phi_m(\altInducingInput_k) \\
  \left[\Kvv\right]_{kk'} & = \langle \psi_{k}, \psi_{k'} \rangle_\cH 
  = \langle \kernel(\altInducingInput_{k}, \argdot), \kernel(\altInducingInput_{k'}, \argdot)) \rangle_\cH 
  = \kernel(\altInducingInput_{k}, \altInducingInput_{k'})
\end{align}
Carrying on from \cref{eq:Qvf-u-spherical-harmonics},
\begin{align}
  \left [ \Qvf \right ]_{kn} 
  & = \sum_{m=1}^M \kernelCoefficient_m \phi_m(\altInducingInput_k) \phi_{m}(\observedInput_n) 
    = \sum_{\level=1}^L \sum_{j=1}^{J_\level^\inputDim} \kernelCoefficient_{\level,j} \sphericalHarmonic{\level}{j}(\altInducingInput_k) \sphericalHarmonic{\level}{j}(\observedInput_n).
\end{align}
Now, as $M \to \infty$ (more precisely, as $L \to \infty$),
$\left [ \Qvf \right ]_{kn}$ converges to $k(\altInducingInput_k, \observedInput_n) = \left[\Kvf\right]_{kn}$. 
In other words, $\left [ \Cvf \right ]_{kn}$ approaches zero in the limit, 
and the orthogonal component of the variational parameterization vanishes,
effectively reducing this combination to standard \gls{SVGP}.
However, for finite $M$, particularly the modest values that are feasible in 
practice, the orthogonal component remains and will strictly improve the 
modelling capacity by capturing the residues that the base component cannot.

\subsubsection{Orthogonal Inducing Variables as Spherical Harmonics} 

Here, orthogonal inducing variables are defined through inter-domain features 
$v_k = \langle f, \psi_k \rangle_\cH$ where $\psi_k \defeq \sphericalHarmonic{\level}{j}$, 
the spherical harmonic of level $\level$ and order $j$, 
and $k$ indexes the pairs $(\level, j)$.
Hence, $\Kvf$ is as expressed in \cref{eq:Kvf},
and
\begin{align}
  \left[\Kvu\right]_{km} & = \langle \psi_k, \phi_m \rangle_\cH = \kernelCoefficient_m^{-1} \delta_{mk} \\
  \left[\Kvv\right]_{kk'} & = \langle \psi_{k}, \psi_{k'} \rangle_\cH 
    = \kernelCoefficient_k^{-1} \delta_{kk'}
\end{align}

This leads to
\begin{equation}
  \left [ \Qvf \right ]_{kn} 
  = \sum_{m=1}^M \kernelCoefficient_m \langle \psi_k, \phi_m \rangle_\cH \phi_{m}(\observedInput_n)
  = \sum_{m=1}^M \delta_{mk} \phi_{m}(\observedInput_n)
  = \phi_k(\observedInput_n)
\end{equation}
in this case, since $\psi_k = \phi_k$, we have $\left [ \Cvf \right ]_{kn} = 0$ 
and this approach collapses to standard \gls{SVGP}, regardless of the value of $M$.

\subsubsection{Orthogonal Inducing Variables as Spherical {NN} Activations} 

Here, orthogonal inducing variables are defined through inter-domain features 
$v_k = \langle f, \psi_k \rangle_\cH$ 
where $\psi_k = H_k$, the $k$-th unit of a 
spherical \gls{NN} activation layer (\cref{eq:spherical-activation-feature}).
Again, $\Kvf$ is as expressed in \cref{eq:Kvf}, and we now have
\begin{align}
  \left[\Kvu\right]_{km} & = \langle \psi_k, \phi_m \rangle_\cH = \kernelCoefficient_m^{-1} \featureCoefficient_m \phi_m(\altInducingInput_k) \\ 
  \left[\Kvv\right]_{kk'} & = \langle \psi_{k}, \psi_{k'} \rangle_\cH  = \sum_{\substack{\level=0: \\ \kernelCoefficient_\level \neq 0}}^{\infty} \frac{\featureCoefficient_\level^2}{\kernelCoefficient_\level} \frac{\level + \alpha}{\alpha} C_{\level}^{(\alpha)} \left ( \frac{\altInducingInput_{k}^\top \altInducingInput_{k'}}{\lVert \altInducingInput_{k} \rVert \lVert \altInducingInput_{k'} \rVert} \right )
\end{align}
which leads to
\begin{equation}
  \left [ \Qvf \right ]_{kn} 
  = \sum_{m=1}^M \featureCoefficient_m \phi_m(\altInducingInput_k) \phi_{m}(\observedInput_n)
\end{equation}
\todo{take care of normalizing constants}
Similar to \cref{ssub:u-spherical-harmonics-v-standard-inducing-points}, 
as $M \to \infty$ (more precisely, as $L \to \infty$),
$\left [ \Qvf \right ]_{kn}$ converges to $\psi_k(\observedInput_n) = \left[\Kvf\right]_{kn}$. 
So again, $\left [ \Cvf \right ]_{kn}$ approaches zero in the limit, 
and the orthogonal component of the variational parameterization vanishes,
effectively reducing this combination to standard \gls{SVGP}.
However, for finite $M$, particularly the small values that are feasible in 
practice, the orthogonal component remains and will strictly improve the 
modelling capacity by capturing the residues that the base component cannot.

\subsection{Base Inducing Variables as Spherical NN Activations}

Here, the base inducing variables are defined through inter-domain features 
$u_m = \langle f, \phi_m \rangle_\cH$ 
where $\phi_m = H_m$, the $m$-th unit of a 
spherical \gls{NN} activation layer (\cref{eq:spherical-activation-feature}).

In this case, $\left[\Kuf\right]_{mn} = \phi_m(\observedInput_{n})$ as in \cref{eq:Kuf}, and
\begin{equation}
  \left[\Kuu\right]_{mm'} 
    = \langle \phi_{m}, \phi_{m'} \rangle_\cH
    = \sum_{\level=0: \kernelCoefficient_\level \neq 0}^{\infty} \frac{\featureCoefficient_\level^2}{\kernelCoefficient_\level} \frac{\level + \alpha}{\alpha} C_{\level}^{(\alpha)} \left ( \frac{\mbz_{m}^\top \mbz_{m'}}{\lVert \mbz_{m} \rVert \lVert \mbz_{m'} \rVert} \right ) 
\end{equation}


\subsubsection{Orthogonal Inducing Variables as Standard Inducing Points}
\label{ssub:orthogonal_inducing_variables_as_standard_inducing_points}

In this case, the orthogonal inducing variables are simply  
standard inducing points $v_k = f(\altInducingInput_m)$, which is equivalent to 
defining $\psi_k: \observedInput \mapsto k(\altInducingInput_k, \observedInput)$.
Hence, we have
\begin{align}
  \left[\Kvf\right]_{kn} 
  & = \psi_k(\observedInput_n) 
    = \kernel(\altInducingInput_{k}, \observedInput_{n}) \\
  \left[\Kvu\right]_{km}
  & = \langle \psi_k, \phi_m \rangle_\cH 
    = \langle \kernel(\altInducingInput_k, \argdot), \phi_m \rangle_\cH
    = \phi_m(\altInducingInput_k) \\
  \left[\Kvv\right]_{kk'} & = \langle \psi_{k}, \psi_{k'} \rangle_\cH 
  = \langle \kernel(\altInducingInput_{k}, \argdot), \kernel(\altInducingInput_{k'}, \argdot)) \rangle_\cH 
  = \kernel(\altInducingInput_{k}, \altInducingInput_{k'})
\end{align}
as in \cref{ssub:u-spherical-harmonics-v-standard-inducing-points}.

\subsubsection{Orthogonal Inducing Variables as Spherical Harmonics} 

Here, orthogonal inducing variables are defined through inter-domain features 
$v_k = \langle f, \psi_k \rangle_\cH$ 
where $\psi_k = H_k$, the $k$-th unit of a 
spherical \gls{NN} activation layer (\cref{eq:spherical-activation-feature}).

We have 
\begin{align}
  \left[\Kvu\right]_{km} & = \langle \psi_k, \phi_m \rangle_\cH = \kernelCoefficient_k^{-1} \featureCoefficient_k \psi_k(\mbz_m) \\
  \left[\Kvf\right]_{kn} & = \langle \psi_k, f(\observedInput_{n}) \rangle_\cH 
    = \langle \psi_k, \kernel(\observedInput_{n}, \argdot) \rangle_\cH  = \psi_k(\observedInput_n) \\
  \left[\Kvv\right]_{kk'} & = \langle \psi_{k}, \psi_{k'} \rangle_\cH 
    = \kernelCoefficient_k^{-1} \delta_{kk'}
\end{align}

This leads to
\begin{equation}
  \left [ \Qvf \right ]_{kn} = \sum_{m=1}^M \sum_{m'=1}^M \left [ \Kuu^{-1} \right ]_{mm'} \kernelCoefficient_k^{-1} \featureCoefficient_k \psi_k(\mbz_m) \phi_{m'}(\observedInput_{n})
\end{equation}

\subsubsection{Orthogonal Inducing Variables as Spherical {NN} Activations} 

Here, orthogonal inducing variables are defined through inter-domain features 
$v_k = \langle f, \psi_k \rangle_\cH$ 
where $\psi_k = H_k$, the $k$-th unit of a 
spherical \gls{NN} activation layer (\cref{eq:spherical-activation-feature}).

In this case, we still get
\begin{align}
  \left[\Kvu\right]_{km} & = \langle \psi_k, \phi_m \rangle_\cH = \sum_{\level=0: \kernelCoefficient_\level \neq 0}^{\infty} \frac{\featureCoefficient_\level^2}{\kernelCoefficient_\level} \frac{\level + \alpha}{\alpha} C_{\level}^{(\alpha)} \left ( \frac{\altInducingInput_{k}^\top \mbz_{m}}{\lVert \altInducingInput_{k} \rVert \lVert \mbz_{m} \rVert} \right ) \\ 
  \left[\Kvf\right]_{kn} & = \langle \psi_k, f(\observedInput_{n}) \rangle_\cH 
    = \langle \psi_k, \kernel(\observedInput_{n}, \argdot) \rangle_\cH  = \psi_k(\observedInput_n) \\
  \left[\Kvv\right]_{kk'} & = \langle \psi_{k}, \psi_{k'} \rangle_\cH  = \sum_{\level=0: \kernelCoefficient_\level \neq 0}^{\infty} \frac{\featureCoefficient_\level^2}{\kernelCoefficient_\level} \frac{\level + \alpha}{\alpha} C_{\level}^{(\alpha)} \left ( \frac{\altInducingInput_{k}^\top \altInducingInput_{k'}}{\lVert \altInducingInput_{k} \rVert \lVert \altInducingInput_{k'} \rVert} \right )
\end{align}

\clearpage 

\section{Experimental Set-up and Implementation Details}
\label{sec:experimental_set_up_and_implementation_details}

\subsection{Hardware}
\label{sub:hardware}

All experiments were carried out on a consumer-grade laptop computer with an 
Intel 
Core\texttrademark{} i7-11800H (8 Cores) @ 4.6GHz Processor, 
16GB Memory, and
a NVIDIA 
GeForce RTX\texttrademark{} 3070 {Laptop (Mobile/Max-Q)} Graphics Card.

\glsunset{LBFGS}

\subsection{Software}
\label{sub:software}

Our method is implemented by extending functionality from the 
\textsf{GPFlow} software library~\cite{GPflow2017}.
The code will be released as open-source software upon publication.
Additional software dependencies upon which our implementation relies, 
either directly or indirectly, are enumerated in \cref{tab:implementations}.

\begin{table}[ht]
\caption{Key software dependencies.}
\label{tab:implementations}
\vskip 0.15in
\begin{center}
\begin{footnotesize}
\begin{tabular}{lll}
\toprule
Method                                       & Software Library             & URL (\texttt{github.com/*})                                                                                                \\
\midrule
\gls{SVGP}~\cite{titsias2009variational}               & \textsf{GPFlow}              & \texttt{\href{https://github.com/GPflow/GPflow}{GPflow/GPflow}}                                                  \\
\gls{ODVGP}~\cite{salimbeni2018orthogonally}           & -                            & \texttt{\href{https://github.com/hughsalimbeni/orth_decoupled_var_gps}{hughsalimbeni/orth\_decoupled\_var\_gps}} \\
\gls{SOLVEGP}~\cite{shi2020sparse}                     & -                            & \texttt{\href{https://github.com/thjashin/solvegp}{thjashin/solvegp}}                                            \\
\textsc{vish}~\cite{dutordoir2020sparse}               & \textsf{Spherical Harmonics} & \texttt{\href{https://github.com/vdutor/SphericalHarmonics}{vdutor/SphericalHarmonics}}                          \\
\textsc{activated} \gls{SVGP}~\cite{dutordoir2021deep} & -                            & \texttt{\href{https://github.com/vdutor/ActivatedDeepGPs}{vdutor/ActivatedDeepGPs}}                              \\
-                                                      & \textsf{Bayesian Benchmarks} & \texttt{\href{https://github.com/hughsalimbeni/bayesian_benchmarks}{hughsalimbeni/bayesian\_benchmarks}}         \\
\bottomrule
\end{tabular}
\end{footnotesize}
\end{center}
\vskip -0.1in
\end{table}

\subsection{Hyperparameters}
\label{sub:hyperparameters}

We adopt sensible defaults across all problems and datasets; no hand-tuning is 
applied to any specific one. 
The choices of the hyperparameters and other relevant dependencies are 
summarized as follows:

\parhead{Optimization.}
We use the \gls{LBFGS} optimizer \cite{byrd1995limited,zhu1997algorithm} with 
the default settings from \texttt{scipy.optimize} \cite{2020SciPy-NMeth}.

\parhead{Likelihood.}
The Gaussian likelihood variance is initialized to 1.0 across all experiments.

\parhead{Kernel parameter initialization.} 
All stationary kernels are initialized with unit lengthscale and amplitude.
\todo{Comment on initialization of the Arccos kernel}

\parhead{Variational parameter initialization.}
The variational distributions $q(\inducingVariables)$, $q(\perpAltInducingVariables)$ 
are initialized with zero mean 
and identity covariance $\variationalLoc = \mbzero, \variationalCov = \mbI$.

\parhead{Whitened parameterization.}
We do \emph{not} use the whitened parameterization (as used, for example, 
by \citet{murray2010slice,hensman2015mcmc}) in 
either $q(\inducingVariables)$ or $q(\perpAltInducingVariables)$.

\parhead{Inducing point initialization.}
We make our best effort to ensure a fair comparison against baselines
involving standard inducing points.
To this end, we adopt the best practice of first optimizing the variational 
parameters, not least the inducing input locations $\mbZ$ (and $\mbW$ where applicable), 
before jointly optimizing all of the free parameters. 
This initialization phase is done for up to 100 iterations of the \gls{LBFGS} algorithm. 

\clearpage

\section{Additional Results}
\label{sec:additional_results}

\subsection{Regression on Airline Delays Dataset}
\label{sub:regression_on_airline_delays_dataset}

\begin{figure}[th]
  \centering
  \includegraphics[width=.7\linewidth]{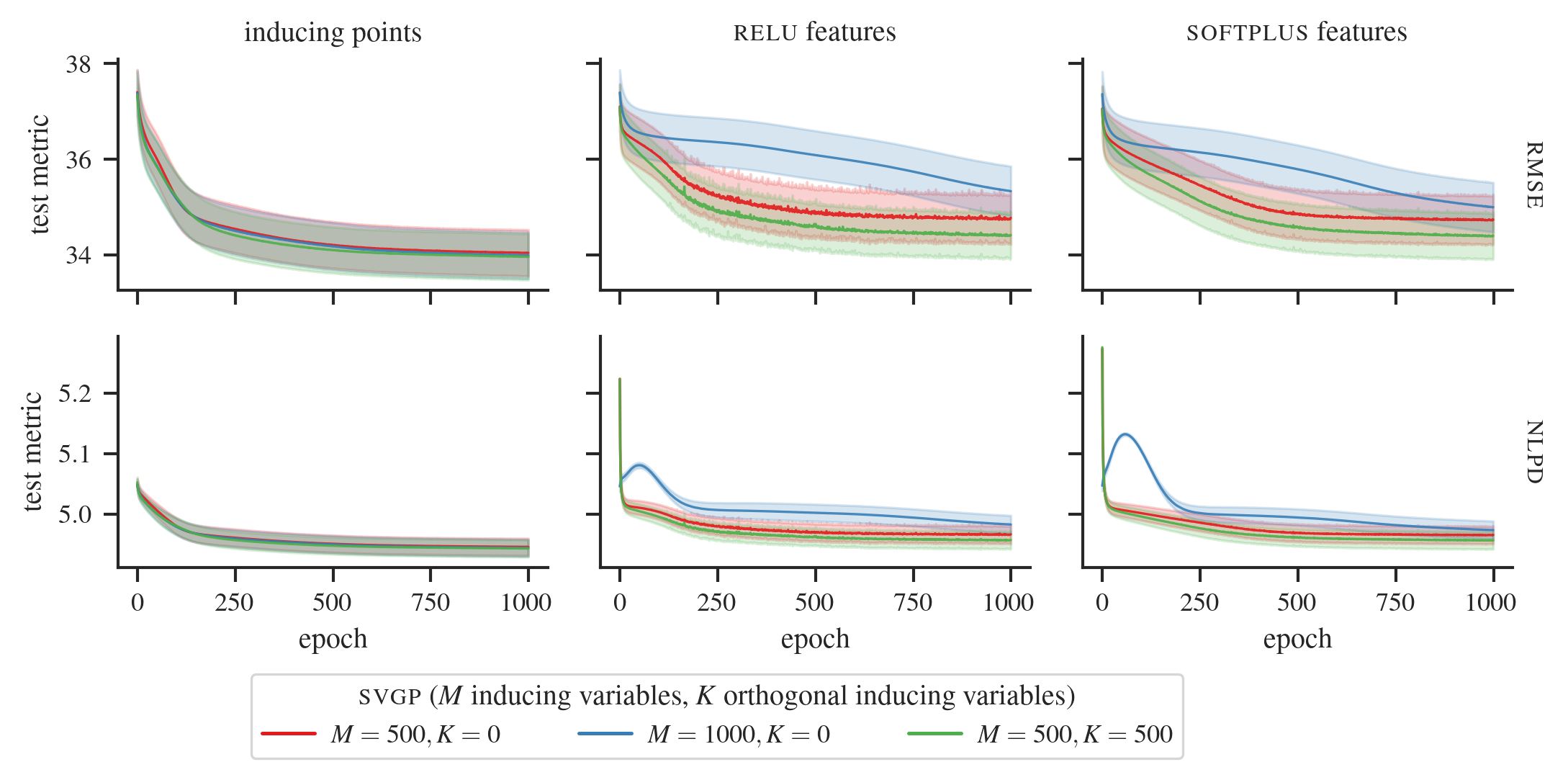}
  \caption{Test metrics, \gls{RMSE} and \gls{NLPD}, 
    aggregated across 5 random sub-sampling test splits
    on a 100K subset of the 2008 U.S. airline delays dataset.
    Results are shown for models using the \emph{Arccos} kernel with 
    standard inducing points and various activation features.
    Along the column labeled ``inducing points'', 
    the red and blue lines (\mplred{---} and \mplblue{---}) 
    represent the mini-batch \gls{SVGP} \cite{hensman2013gaussian},
    while the green line (\mplgreen{---}) 
    represents \gls{SOLVEGP} \cite{shi2020sparse}.
    Along the column labeled ``\gls{SOFTPLUS} features'', 
    the red and blue lines (\mplred{---} and \mplblue{---}) 
    represent the \textsc{activated} \gls{SVGP} \cite{dutordoir2021deep},
    while the green line (\mplgreen{---}) represents our proposed approach.
  }
  \label{fig:airline-regression-100k}
\end{figure}

We repeat the experiment outlined in \cref{sub:large_scale_regression_on_airline_delays_dataset},
focusing on a reduced subset of the 2008 U.S. airline delays dataset that 
consists of 100K randomly selected observations. 
Unlike the previous experimental set-up, the parameters are optimised for a 
total of 1,000 epochs.
Additionally, we report aggregated results from 5 repetitions
across $\nicefrac{1}{3}$ held-out test sets.
The results are shown in~\cref{fig:airline-regression-100k}.

\subsection{Extra UCI Repository Datasets}
\label{sub:results_on_larger_uci_repository_datasets}

\begin{figure*}[ht]
  \vskip 0.2in
  \begin{center}
  \centerline{\includegraphics[width=\linewidth]{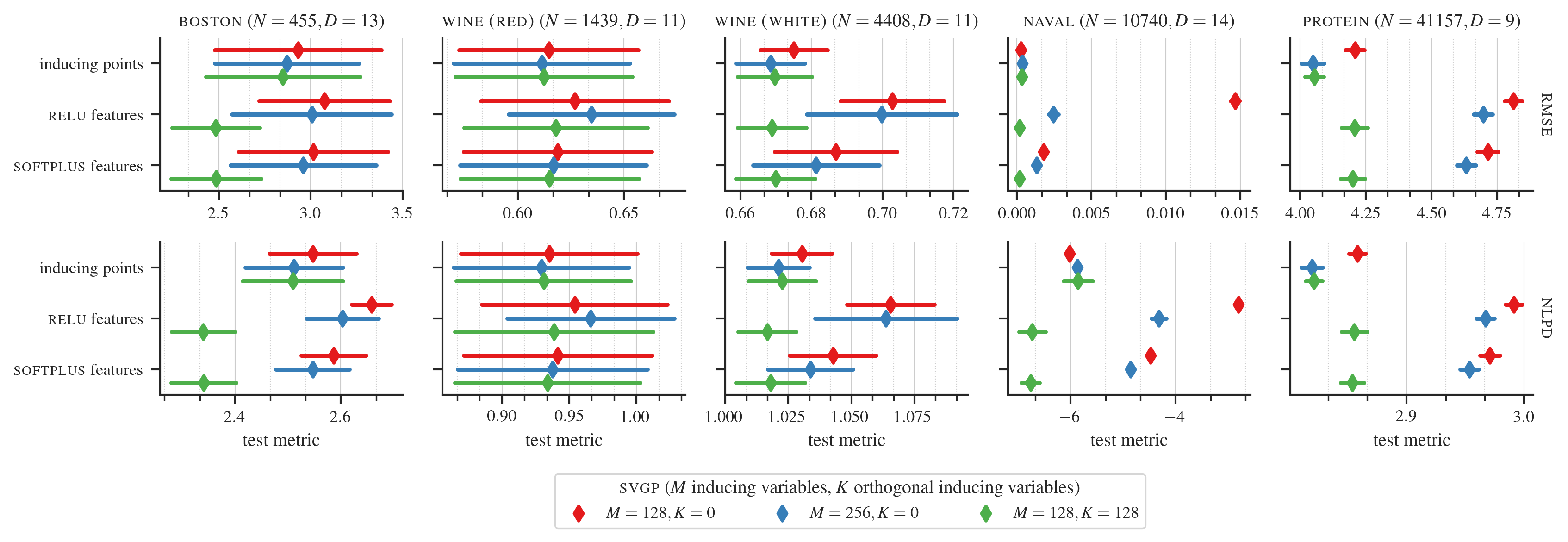}}
  \caption{Test metrics, \gls{RMSE} and \gls{NLPD}, on an extra set of larger 
    regression problems from the \gls{UCI} dataset repository for 
    the \emph{Arccos} kernel and various activation features.
    Along the rows labeled \emph{``inducing points''}, 
    the red and blue markers ($\mplred{\blacklozenge}, \mplblue{\blacklozenge}$) 
    represent the original \gls{SVGP} model \cite{titsias2009variational},
    while the green markers ($\mplgreen{\blacklozenge}$) 
    represent \gls{SOLVEGP} \cite{shi2020sparse}.
    Along the remaining rows, 
    the red and blue markers ($\mplred{\blacklozenge}, \mplblue{\blacklozenge}$) 
    represent the \textsc{activated} \gls{SVGP} \cite{dutordoir2021deep},
    while the green markers ($\mplgreen{\blacklozenge}$) represent our 
    proposed approach.
  }
  \label{fig:uci-regression-extra}
  \end{center}
  \vskip -0.2in
\end{figure*}

Results on a few larger regression datasets from the \gls{UCI} repository can 
be found in \cref{fig:uci-regression-extra}.
In this analysis, we adopted the same combination of activation features 
and sparse \gls{GP} models as described in \cref{sub:regression_on_uci_repository_datasets}. 
However, in contrast to \cref{sub:regression_on_uci_repository_datasets}, 
we restrict our focus to the Arccos kernel.

\subsection{Numerical Tables}
\label{sub:numerical_tables}

For completeness, we include the numerical values of 
underlying \cref{fig:uci-regression-rmse,fig:uci-regression-nlpd} 
from \cref{sec:experiments} in \cref{tab:uci-rmse,tab:uci-nlpd} below.

\begin{table*}[ht]
\caption{Numerical values of the test \gls{RMSE} corresponding to \cref{fig:uci-regression-rmse}.} 
\label{tab:uci-rmse}
\vskip 0.15in
\begin{center}
\begin{tiny}
\begin{tabular}{lp{60pt}cccccc}
\toprule
                              & inducing variable & \multicolumn{2}{c}{\textsc{relu} features} & \multicolumn{2}{c}{\textsc{softplus} features} & \multicolumn{2}{c}{inducing points} \\
                              & kernel &                 Arccos & Mat\'{e}rn-$\nicefrac{5}{2}$ &                     Arccos & Mat\'{e}rn-$\nicefrac{5}{2}$ &          Arccos & Mat\'{e}rn-$\nicefrac{5}{2}$ \\
dataset & \textsc{svgp} ($M$ inducing variables, $K$ orthogonal inducing variables) &                        &                              &                            &                              &                 &                              \\
\midrule
\multirow{3}{*}{\textsc{concrete} ($N=927, D=8$)} & $M=128, K=0$ &          $6.82\pm0.30$ &                $6.92\pm0.25$ &              $6.69\pm0.29$ &                $6.89\pm0.26$ &   $6.58\pm0.37$ &                $5.94\pm0.25$ \\
                              & $M=128, K=128$ &          $5.93\pm0.38$ &                $5.87\pm0.30$ &              $6.06\pm0.32$ &                $5.91\pm0.33$ &   $6.40\pm0.40$ &                $5.75\pm0.35$ \\
                              & $M=256, K=0$ &          $6.56\pm0.20$ &                $6.87\pm0.26$ &              $6.45\pm0.24$ &                $6.86\pm0.26$ &   $6.40\pm0.35$ &                $5.64\pm0.29$ \\
\cline{1-8}
\multirow{3}{*}{\textsc{energy} ($N=691, D=8$)} & $M=128, K=0$ &          $1.36\pm0.30$ &                $1.58\pm0.17$ &              $0.95\pm0.10$ &                $1.53\pm0.16$ &   $0.47\pm0.08$ &                $0.47\pm0.08$ \\
                              & $M=128, K=128$ &          $0.47\pm0.08$ &                $0.47\pm0.08$ &              $0.47\pm0.08$ &                $0.47\pm0.08$ &   $0.48\pm0.08$ &                $0.47\pm0.08$ \\
                              & $M=256, K=0$ &          $0.91\pm0.08$ &                $1.51\pm0.15$ &              $0.86\pm0.08$ &                $1.51\pm0.15$ &   $0.47\pm0.08$ &                $0.46\pm0.08$ \\
\cline{1-8}
\multirow{3}{*}{\textsc{kin8nm} ($N=7372, D=8$)} & $M=128, K=0$ &          $0.12\pm0.00$ &                $0.10\pm0.00$ &              $0.10\pm0.00$ &                $0.10\pm0.00$ &   $0.10\pm0.00$ &                $0.08\pm0.00$ \\
                              & $M=128, K=128$ &          $0.08\pm0.00$ &                $0.08\pm0.00$ &              $0.08\pm0.00$ &                $0.08\pm0.00$ &   $0.09\pm0.00$ &                $0.08\pm0.00$ \\
                              & $M=256, K=0$ &          $0.10\pm0.00$ &                $0.10\pm0.00$ &              $0.09\pm0.00$ &                $0.10\pm0.00$ &   $0.09\pm0.00$ &                $0.08\pm0.00$ \\
\cline{1-8}
\multirow{3}{*}{\textsc{power} ($N=8611, D=4$)} & $M=128, K=0$ &          $4.20\pm0.16$ &                $4.22\pm0.15$ &              $4.20\pm0.16$ &                $4.22\pm0.15$ &   $3.93\pm0.18$ &                $3.90\pm0.19$ \\
                              & $M=128, K=128$ &          $3.96\pm0.18$ &                $3.96\pm0.19$ &              $3.91\pm0.19$ &                $3.96\pm0.19$ &   $3.85\pm0.18$ &                $3.71\pm0.19$ \\
                              & $M=256, K=0$ &          $4.20\pm0.16$ &                $4.22\pm0.15$ &              $4.20\pm0.16$ &                $4.22\pm0.15$ &   $3.83\pm0.19$ &                $3.68\pm0.19$ \\
\cline{1-8}
\multirow{3}{*}{\textsc{yacht} ($N=277, D=6$)} & $M=128, K=0$ &          $1.17\pm0.36$ &                $2.43\pm0.54$ &              $1.16\pm0.36$ &                $2.43\pm0.54$ &   $1.06\pm0.24$ &                $0.34\pm0.19$ \\
                              & $M=128, K=128$ &          $0.59\pm0.23$ &                $0.51\pm0.25$ &              $0.60\pm0.23$ &                $0.49\pm0.29$ &   $1.06\pm0.23$ &                $0.34\pm0.19$ \\
                              & $M=256, K=0$ &          $1.16\pm0.35$ &                $2.43\pm0.54$ &              $1.13\pm0.36$ &                $2.43\pm0.54$ &   $1.07\pm0.23$ &                $0.28\pm0.14$ \\
\bottomrule
\end{tabular}

\end{tiny}
\end{center}
\vskip -0.1in
\end{table*}

\begin{table*}[ht]
\caption{Numerical values of the test \gls{NLPD} corresponding to \cref{fig:uci-regression-nlpd}.} 
\label{tab:uci-nlpd}
\vskip 0.15in
\begin{center}
\begin{tiny}
\begin{tabular}{lp{60pt}cccccc}
\toprule
                              & inducing variable & \multicolumn{2}{c}{\textsc{relu} features} & \multicolumn{2}{c}{\textsc{softplus} features} & \multicolumn{2}{c}{inducing points} \\
                              & kernel &                 Arccos & Mat\'{e}rn-$\nicefrac{5}{2}$ &                     Arccos & Mat\'{e}rn-$\nicefrac{5}{2}$ &          Arccos & Mat\'{e}rn-$\nicefrac{5}{2}$ \\
dataset & \textsc{svgp} ($M$ inducing variables, $K$ orthogonal inducing variables) &                        &                              &                            &                              &                 &                              \\
\midrule
\multirow{3}{*}{\textsc{concrete} ($N=927, D=8$)} & $M=128, K=0$ &          $3.33\pm0.04$ &                $3.36\pm0.02$ &              $3.31\pm0.04$ &                $3.36\pm0.03$ &   $3.30\pm0.05$ &                $3.19\pm0.07$ \\
                              & $M=128, K=128$ &          $3.19\pm0.08$ &                $3.18\pm0.08$ &              $3.22\pm0.08$ &                $3.18\pm0.08$ &   $3.27\pm0.06$ &                $3.12\pm0.08$ \\
                              & $M=256, K=0$ &          $3.29\pm0.03$ &                $3.35\pm0.03$ &              $3.27\pm0.04$ &                $3.35\pm0.03$ &   $3.26\pm0.05$ &                $3.11\pm0.09$ \\
\cline{1-8}
\multirow{3}{*}{\textsc{energy} ($N=691, D=8$)} & $M=128, K=0$ &          $1.83\pm0.12$ &                $2.03\pm0.04$ &              $1.49\pm0.04$ &                $2.01\pm0.04$ &   $0.70\pm0.17$ &                $0.68\pm0.20$ \\
                              & $M=128, K=128$ &          $0.68\pm0.20$ &                $0.68\pm0.20$ &              $0.69\pm0.20$ &                $0.69\pm0.19$ &   $0.70\pm0.18$ &                $0.67\pm0.19$ \\
                              & $M=256, K=0$ &          $1.43\pm0.03$ &                $1.99\pm0.04$ &              $1.38\pm0.03$ &                $1.99\pm0.04$ &   $0.70\pm0.18$ &                $0.67\pm0.20$ \\
\cline{1-8}
\multirow{3}{*}{\textsc{kin8nm} ($N=7372, D=8$)} & $M=128, K=0$ &         $-0.69\pm0.02$ &               $-0.78\pm0.02$ &             $-0.83\pm0.02$ &               $-0.82\pm0.02$ &  $-0.85\pm0.02$ &               $-1.03\pm0.02$ \\
                              & $M=128, K=128$ &         $-1.03\pm0.02$ &               $-1.03\pm0.02$ &             $-1.03\pm0.02$ &               $-1.03\pm0.02$ &  $-0.90\pm0.01$ &               $-1.10\pm0.02$ \\
                              & $M=256, K=0$ &         $-0.89\pm0.02$ &               $-0.83\pm0.02$ &             $-0.95\pm0.02$ &               $-0.84\pm0.02$ &  $-0.90\pm0.01$ &               $-1.10\pm0.02$ \\
\cline{1-8}
\multirow{3}{*}{\textsc{power} ($N=8611, D=4$)} & $M=128, K=0$ &          $2.85\pm0.04$ &                $2.86\pm0.03$ &              $2.85\pm0.04$ &                $2.86\pm0.03$ &   $2.79\pm0.04$ &                $2.78\pm0.05$ \\
                              & $M=128, K=128$ &          $2.80\pm0.05$ &                $2.80\pm0.05$ &              $2.79\pm0.05$ &                $2.80\pm0.05$ &   $2.77\pm0.04$ &                $2.73\pm0.05$ \\
                              & $M=256, K=0$ &          $2.85\pm0.04$ &                $2.86\pm0.03$ &              $2.85\pm0.04$ &                $2.86\pm0.03$ &   $2.77\pm0.05$ &                $2.73\pm0.05$ \\
\cline{1-8}
\multirow{3}{*}{\textsc{yacht} ($N=277, D=6$)} & $M=128, K=0$ &          $1.81\pm0.08$ &                $2.51\pm0.06$ &              $1.80\pm0.08$ &                $2.51\pm0.06$ &   $1.50\pm0.15$ &                $0.33\pm0.25$ \\
                              & $M=128, K=128$ &          $0.91\pm0.26$ &                $0.73\pm0.40$ &              $0.92\pm0.26$ &                $0.70\pm0.41$ &   $1.51\pm0.17$ &                $0.30\pm0.23$ \\
                              & $M=256, K=0$ &          $1.81\pm0.08$ &                $2.51\pm0.06$ &              $1.79\pm0.08$ &                $2.51\pm0.06$ &   $1.52\pm0.18$ &                $0.07\pm0.30$ \\
\bottomrule
\end{tabular}

\end{tiny}
\end{center}
\vskip -0.1in
\end{table*}

\end{document}